\documentclass[journal]{IEEEtran}
\usepackage{amsmath,epsfig,bm}
\usepackage{multirow}
\usepackage{balance}
\usepackage{hyperref}
\usepackage{cite}
\usepackage{graphicx}
\usepackage{svg}
\usepackage{lipsum}
\usepackage{url}
\graphicspath{{Figures/}}
\usepackage{caption}
\usepackage{subcaption} %should come after caption package if given, can't be used with caption2, defines subfloat
\usepackage{color}
\usepackage{balance}
\usepackage{float}
\usepackage{multirow}
\usepackage{float}
\usepackage{breqn}
\usepackage{tabularx}
\usepackage{comment}
\usepackage{censor}
\usepackage{wrapfig}
\usepackage{amssymb}%checkmark
\usepackage{ragged2e}
\usepackage{gensymb}
\usepackage{array}
\usepackage{booktabs}
\usepackage{threeparttable}
\usepackage{wasysym}
\usepackage{algorithm}
\usepackage{algpseudocode}
\definecolor{CodeGray}{rgb}{0.97,0.97,0.97}
\usepackage{listings} % syntax highlighting for sparql
\lstdefinelanguage{SPARQL}{
  basicstyle=\small\ttfamily,
  backgroundcolor=\color{CodeGray},
  columns=fullflexible,
  breaklines=true,
  sensitive=true,
  frame=single,
  aboveskip=1em,
  belowskip=1em,
  xleftmargin=.5em,
  xrightmargin=.5em,
  framexleftmargin=.5em,
  framextopmargin=.5em,
  framexbottommargin=.5em,
  framexrightmargin=.5em,
  tabsize = 2,
  showstringspaces=false,
  morecomment=[l][\color{gray}]{\#},       % comments
  morecomment=[n][\color{blue}]{<http}{>}, % uris
  morestring=[b][\color{OliveGreen}]{\"},  % strings
  keywordsprefix=?,
  classoffset=0,
  keywordstyle=\color{Sepia},
  morekeywords={},
  classoffset=1,
  keywordstyle=\color{Green},
  morekeywords={ca, seas,rdf,rdfs,owl,xsd,purl, qudt, prov, sosa, ca_prop, ca_type,geo, geos, car, geof, spatialF, lgdo, unit},
  classoffset=2,
  keywordstyle=\color{MidnightBlue},
  morekeywords={
    SELECT,CONSTRUCT,DESCRIBE,ASK,WHERE,FROM,NAMED,PREFIX,BASE,OPTIONAL,FILTER,GRAPH,LIMIT,OFFSET,SERVICE,UNION,EXISTS,NOT,BINDINGS,MINUS,a, ORDER,BY, GROUP, AS, VALUES, BIND, STRDT, CONCAT, STR
  }
}

\usepackage[colorinlistoftodos,prependcaption,textsize=tiny]{todonotes}
\usepackage{nomencl}
\makenomenclature

\begin{document}

\title{A Deep Learning Approach for Spatio-Temporal Forecasting of InSAR Ground Deformation in Eastern Ireland}

%\begin{comment}
\author{Wendong~Yao, Saeed~Azadnejad, Binhua Huang, Shane~Donohue, and~Soumyabrata~Dev,~\IEEEmembership{Senior~Member,~IEEE}% <-this % stops a space
\thanks{Manuscript received September XX, 2025; revised September XX, 2025; accepted September XX, 2025.}
\thanks{In the spirit of reproducible research, all the code and dataset used in this paper can be accessed here: \href{https://github.com/WendongYao/Wendong_2025_Spatiotemporal_InSAR_dataset_prediction}{Wendong_2025_Spatiotemporal_InSAR_dataset_prediction}}
\thanks{W.\ Yao and S.\ Dev are with ADAPT SFI Research Centre, School of Computer Science, University College Dublin, Ireland.}
\thanks{S. Azadnejad and S.\ Donohue are with the School of Civil Engineering, University College Dublin, Ireland.}
\thanks{Send correspondence to S.\ Dev, e-mail: soumyabrata.dev@ucd.ie.}
}
%\end{comment}

\markboth{IEEE Transactions on Geoscience and Remote Sensing,~Vol.~XX, No.~XX, XXX~2025}%
{Shell \MakeLowercase{\textit{et al.}}: Bare Demo of IEEEtran.cls for IEEE Journals}


\maketitle

\begin{abstract}
Monitoring ground displacement is crucial for urban infrastructure stability and mitigating geological hazards. However, forecasting future deformation from sparse Interferometric Synthetic Aperture Radar (InSAR) time-series data remains a significant challenge. This paper introduces a novel deep learning framework that transforms these sparse point measurements into a dense spatio-temporal tensor. This methodological shift allows, for the first time, the direct application of advanced computer vision architectures to this forecasting problem. We design and implement a hybrid Convolutional Neural Network and Long-Short Term Memory (CNN-LSTM) model, specifically engineered to simultaneously learn spatial patterns and temporal dependencies from the generated data tensor. The model's performance is benchmarked against powerful machine learning baselines, Light Gradient Boosting Machine and LASSO regression, using Sentinel-1 data from eastern Ireland. Results demonstrate that the proposed architecture provides significantly more accurate and spatially coherent forecasts, establishing a new performance benchmark for this task. Furthermore, an interpretability analysis reveals that baseline models often default to simplistic persistence patterns, highlighting the necessity of our integrated spatio-temporal approach to capture the complex dynamics of ground deformation. Our findings confirm the efficacy and potential of spatio-temporal deep learning for high-resolution deformation forecasting.

\end{abstract}


% \fi
%
% For peer review papers, this IEEEtran command inserts a page break and
\IEEEpeerreviewmaketitle

\section{Introduction}
\label{sec:intro} 
Ground movement in rapidly growing urban areas is a critical concern because it can compromise the integrity of civil infrastructure and buildings, posing significant risks to public safety and economic stability. While this movement can be triggered by various factors, significant ground deformation is often linked to subsurface changes, such as groundwater extraction or tunneling, which can induce widespread subsidence. In other cases, localized settlement may result from large-scale construction, particularly on soft, compressible soils. Given these potential impacts, monitoring and accurately forecasting ground deformation is essential for urban planning, risk mitigation, and ensuring the long-term resilience of critical infrastructure. This highlights the need to monitor and analyze ground motion in urban settings \cite{GroundWater}. As noted in \cite{DisAndLandS}, monitoring ground displacement is vital for evaluating landslides, which occur worldwide and can pose a considerable risk to life and critical infrastructure \cite{landDis}. Furthermore, such deformation affects energy systems, as stable ground conditions are crucial to the safe and efficient function of key energy infrastructures, such as power plants, energy storage systems, and distribution networks. Numerous studies have concentrated on the analysis of deformation in urban areas, including efforts to understand the effect of movement of the ground on the planning, operation and resilience of the energy infrastructure\cite{UrbanDef1,UrbanDef2,UrbanDef3}.

Various techniques have been used to monitor ground displacement, with interferometric synthetic aperture radar (InSAR) systems delivering significant results \cite{OSMANOGLU201690}. This system facilitates the creation of precise time series data for ground deformation \cite{EGMSDescript,AZADNEJAD2020101950}. In the realm of remote sensing, machine learning technology is widely applied and has demonstrated substantial effectiveness in recognition and estimation processing tasks \cite{EuroSATYao}, \cite{AZEEM2024100515}. Numerous studies have explored time series estimation of ground displacement, predominantly using machine learning models on datasets generated by InSAR, such as in \cite{InSARLS1} and \cite{InSARLS2}. However, fewer investigations have focused on the spatial estimation of derived InSAR data sets.

While existing studies have successfully applied machine learning to InSAR datasets, they often treat the prediction as a purely temporal or spatial problem, failing to fully capture the complex interplay between the two domains. In a foundational study\cite{WesIe}, we explored a purely spatial prediction approach, establishing a methodological framework for feature evaluation. However, that analysis revealed a key limitation: while certain spatial features like acceleration were significant, the model's performance was ultimately constrained by the absence of temporal dynamics, achieving a modest estimation accuracy of 69.7\%. This finding underscored that a more accurate forecast requires the integration of both spatial patterns and their temporal evolution. This gap motivates our current work, where we develop a dedicated spatio-temporal deep learning model to overcome the limitations of a purely spatial approach.

To achieve this, we introduce a new methodology founded on a regenerated spatio-temporal dataset and a bespoke hybrid deep learning model. This paper details this approach and its evaluation. The main contributions are:

 \begin{enumerate}
    \item \textbf{Methodological Innovation for InSAR Time-Series Analysis:} We introduce a novel end-to-end framework that transforms sparse, point-based InSAR measurements into a dense spatio-temporal tensor format. This methodological shift enables, for the first time in this context, the direct application of advanced spatio-temporal deep learning architectures for ground deformation forecasting, moving beyond traditional point-wise or purely spatial analyses.
    \item \textbf{Demonstration of a Hybrid Deep Learning Architecture for Spatio-Temporal Forecasting:} We design and validate a hybrid CNN-LSTM model, demonstrating the power of architectures that explicitly and simultaneously learn both spatial patterns and temporal dependencies from InSAR data. Our results show that this integrated approach significantly outperforms powerful machine learning baselines that do not model the spatio-temporal structure, establishing a new performance benchmark for this prediction task.
    \item \textbf{Enhancing Model Interpretability in a Geospatial Context:} We apply SHAP analysis to provide insights into the decision-making process of a high-performing baseline model (LightGBM). This analysis not only reveals the limitations of simpler autoregressive patterns but also highlights the necessity of the more complex feature learning captured by our proposed deep learning model, thereby contributing to the broader effort of building more transparent and trustworthy AI models in the remote sensing domain.
\end{enumerate}

\section{Study Area and Dataset Description}
\label{S2}
\begin{figure*}[htb]
  \centering
  \includegraphics[scale=0.9]{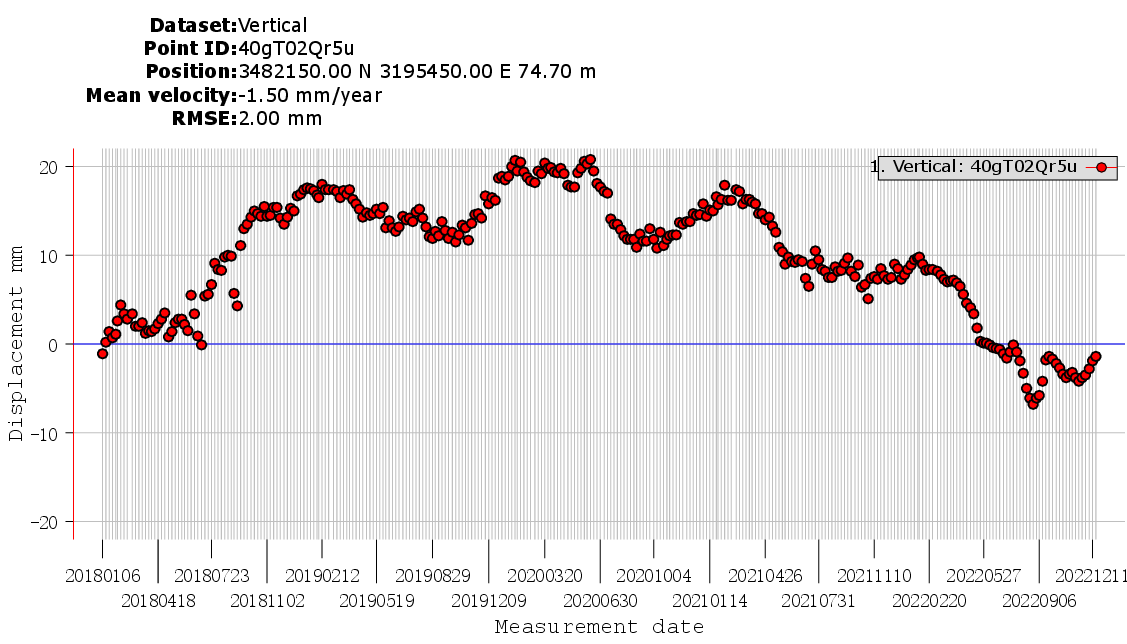}
  \caption{Example of how EGMS presents data for a single geographical point\cite{EGMS}}
  \label{fig:plot_eps}
\end{figure*}

Our study focuses on an area in eastern Ireland, specifically at the coordinates Easting = 32 and Northing = 34. We utilized the  European Ground Motion Service (EGMS) 'Level-3 2018-2022' vector dataset, which provides annual updates throughout Europe. This dataset is derived from Copernicus Sentinel-1 satellite data and measures ground movements, incorporating results within the data set. Data generation employs persistent and distributed scatterer radar interferometry for time series analysis\cite{EGMS}. Figure.~\ref{fig:plot_eps} illustrates data from a particular geographical location, capturing essential parameters such as velocity, root mean square error (RMSE), and displacement.

The dataset we obtained provides ground displacement values in millimeters for each geographic point it records. The locations are logged using the ETRS89-LAEA standard\cite{EGMSDescript}. For our study area, coordinates are captured at 100-meter intervals, covering shifts both eastward and northward. The displacement variation of each point is tracked. From January 2018 to December 2022, displacement values for each point were recorded every 6 days.

The dataset is used by taking the final displacement value recorded as the target. The model is trained using time series data on the ground displacement at a specific location and produces an estimate of the displacement value on the last recorded day.

We used an Ortho-leveled dataset from EGMS, which includes measurements of vertical and horizontal displacements, specifically from East to West. These measurements are derived from calibrated data and resampled onto a regular grid with cells measuring 100 by 100 meters \cite{EGMSDescript}. Our study uses data from 2018 to 2022.\footnote{The EGMS dataset is available for download at: \url{https://egms.land.copernicus.eu/}}.

\section{Methodology}
\begin{figure*}[htbp]
  \centerline{\includegraphics[scale=0.5]{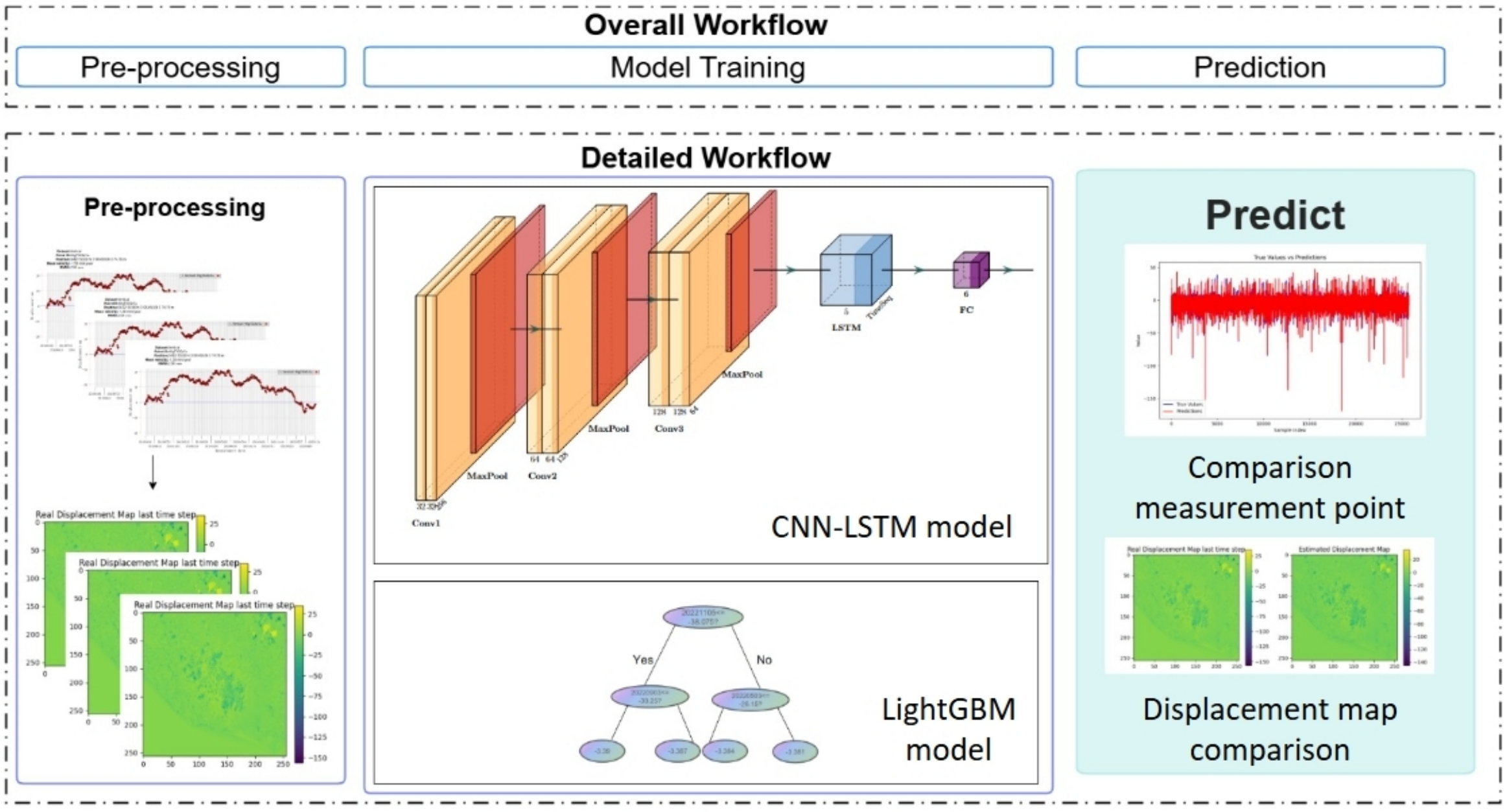}}
  \caption{Workflow of our methodology, including data generation, model training and performance assessment}
  \label{fig:workflow}
  \end{figure*}

In this section, we will introduce the methodology we used to generate our dataset and the machine learning models we applied to it. A detailed graph of workflow is shown in Figure.~\ref{fig:workflow}. Details of the workflow, including the data generation process, the machine learning model training process, and the performance assessment process, will be discussed in the following subsections.

Our training data are generated from the EGMS dataset into maps for displacement. Detailed data generation process will be introduced in Section.~\ref{DataGeneration}.

For model training and assessment, we generated a CNN-LSTM model with spatiotemporal attention mechanisms. We also used the LightGBM model, which also showed good
performance on similar tasks\cite{LightGBM}, for comparison. The training and testing
pipeline framework will be introduced in Section.~\ref{MLIntro}.

For performance comparison, we used parameters similar to those in previous research, to compare our models' performance. Scatter plot, residual plot and residual boxplot are generated to clearly reveal the performance of the models. We also used the results to generate a "displacement map", showing the predicted deformation in the region. We also compared the predicted displacement map and the actual displacement map to clearly see the difference. Performance comparison will be introduced in Section.~\ref{PerformanceComparison}.

Algorithm.~\ref{alg:main_pipeline} shows in detail how the experiment is performed, and Table.~\ref{tab:main_notation} describes the variables shown in the algorithm.

\begin{algorithm*}[htbp]
\caption{Spatio-Temporal InSAR Deformation Forecasting and Analysis Pipeline}
\label{alg:main_pipeline}
\begin{algorithmic}[1]
\small
\Require
    $\mathcal{D}_{\text{raw}}$: Raw Sentinel-1 measurement data as a CSV file, containing columns for Easting, Northing, and displacement time series. \\
    $T_{\text{in}} \leftarrow 300$: Number of input time steps for training (columns 11 to 311). \\
    $T_{\text{out}}$: The target time step for prediction (column 313). \\
    $H, W \leftarrow 256, 256$: The spatial resolution of the output grid. \\
    $\mathcal{M} \leftarrow \{\mathcal{M}_{\text{CNN-LSTM}}, \mathcal{M}_{\text{LGBM}}, \mathcal{M}_{\text{Lasso}}\}$: Set of machine learning models for comparison.

\Ensure
    $\{\hat{\mathbf{Y}}_m\}_{m \in \mathcal{M}}$: Predicted displacement maps for each model. \\
    $\{\text{RMSE}_m, \text{R}^2_m\}_{m \in \mathcal{M}}$: Performance metrics (RMSE, R²) for each model. \\
    $\mathbf{\Phi}_{\text{LGBM}}$: SHAP values for interpreting the LightGBM model.

\State \textbf{Part 1: Spatio-Temporal Data Gridding}
\State Let $\mathbf{p} \leftarrow (\text{Easting}, \text{Northing})$ be the coordinates of measurement points from $\mathcal{D}_{\text{raw}}$.
\State Define a regular grid $\mathcal{G}$ of size $H \times W$ spanning the extent of $\mathbf{p}$.
\State Initialize an empty tensor for input features $\mathcal{X} \in \mathbb{R}^{T_{\text{in}} \times H \times W}$.
\For{$t = 1, \dots, T_{\text{in}}$}
    \State Let $\mathbf{v}_t$ be the displacement values at time $t$ from $\mathcal{D}_{\text{raw}}$.
    \State Interpolate sparse data $(\mathbf{p}, \mathbf{v}_t)$ onto the grid $\mathcal{G}$ to create a dense map $\mathbf{X}_t$. \Comment{Using `scipy.interpolate.griddata` with 'linear' method}
    \State Handle missing values in $\mathbf{X}_t$ by replacing NaN with zero.
    \State $\mathcal{X}[t-1, :, :] \leftarrow \mathbf{X}_t$.
\EndFor
\State Let $\mathbf{v}_{\text{target}}$ be the displacement values at time $T_{\text{out}}$.
\State Interpolate $(\mathbf{p}, \mathbf{v}_{\text{target}})$ onto $\mathcal{G}$ to create the ground truth map $\mathbf{Y}_{\text{true}} \in \mathbb{R}^{H \times W}$.

\Statex
\State \textbf{Part 2: Model Training and Prediction}
\For{each model $\mathcal{M}_i$ in $\mathcal{M}$}
    \If{$\mathcal{M}_i$ is $\mathcal{M}_{\text{CNN-LSTM}}$}
        \State Reshape input to $\mathcal{X}_{\text{5D}} \in \mathbb{R}^{1 \times T_{\text{in}} \times 1 \times H \times W}$. \Comment{Batch, Time, Channel, Height, Width}
        \State Train $\mathcal{M}_{\text{CNN-LSTM}}$ on $(\mathcal{X}_{\text{5D}}, \mathbf{Y}_{\text{true}})$ using MSE loss.
        \State Predict $\hat{\mathbf{Y}}_{\text{CNN-LSTM}} \leftarrow \mathcal{M}_{\text{CNN-LSTM}}(\mathcal{X}_{\text{5D}})$.
    \ElsIf{$\mathcal{M}_i$ is $\mathcal{M}_{\text{LGBM}}$ or $\mathcal{M}_{\text{Lasso}}$}
        \State Reshape input $\mathcal{X}$ to a tabular matrix $\mathbf{X}_{\text{table}} \in \mathbb{R}^{(H \cdot W) \times T_{\text{in}}}$. \Comment{Each row is a pixel's time series}
        \State Reshape target $\mathbf{Y}_{\text{true}}$ to a vector $\mathbf{y}_{\text{true}} \in \mathbb{R}^{H \cdot W}$.
        \If{$\mathcal{M}_i$ is $\mathcal{M}_{\text{LGBM}}$}
            \State Split $(\mathbf{X}_{\text{table}}, \mathbf{y}_{\text{true}})$ into training and validation sets.
            \State Train $\mathcal{M}_{\text{LGBM}}$ with early stopping.
            \State Predict on the full dataset: $\hat{\mathbf{y}}_{\text{LGBM}} \leftarrow \mathcal{M}_{\text{LGBM}}(\mathbf{X}_{\text{table}})$.
            \State Reshape prediction back to a map: $\hat{\mathbf{Y}}_{\text{LGBM}} \in \mathbb{R}^{H \times W}$.
        \ElsIf{$\mathcal{M}_i$ is $\mathcal{M}_{\text{Lasso}}$}
            \State Standardize features $\mathbf{X}_{\text{table}}$ and target $\mathbf{y}_{\text{true}}$.
            \State Train $\mathcal{M}_{\text{Lasso}}$ by optimizing MSE + L1 regularization loss.
            \State Predict on the full dataset and inverse-transform to get $\hat{\mathbf{y}}_{\text{Lasso}}$.
            \State Reshape prediction back to a map: $\hat{\mathbf{Y}}_{\text{Lasso}} \in \mathbb{R}^{H \times W}$.
        \EndIf
    \EndIf
\EndFor

\Statex
\State \textbf{Part 3: Performance Assessment and Interpretation}
\For{each model prediction $\hat{\mathbf{Y}}_m$}
    \State Compute pixel-wise metrics: $\text{RMSE}_m \leftarrow \sqrt{\text{MSE}(\mathbf{Y}_{\text{true}}, \hat{\mathbf{Y}}_m)}$.
    \State Compute coefficient of determination: $\text{R}^2_m \leftarrow \text{R}^2(\mathbf{Y}_{\text{true}}, \hat{\mathbf{Y}}_m)$.
    \State Generate and save comparison plots of $\mathbf{Y}_{\text{true}}$ versus $\hat{\mathbf{Y}}_m$.
\EndFor
\State \textbf{SHAP Analysis for LightGBM}
\State Initialize explainer: $\mathcal{E} \leftarrow \text{shap.TreeExplainer}(\mathcal{M}_{\text{LGBM}})$.
\State Compute SHAP values $\mathbf{\Phi}_{\text{LGBM}}$ on a subset of the test data.
\State Generate and save global feature importance summary plot.
\State Generate and save local prediction force plot for a sample pixel.
\end{algorithmic}
\end{algorithm*}

\begin{table}[htbp]
\centering
\small
\begin{tabular}{p{2.5cm} p{5cm}}
\toprule
\textbf{Symbol} & \textbf{Description} \\
\midrule
\multicolumn{2}{l}{\textbf{Input Data \& Parameters}} \\
$\mathcal{D}_{\text{raw}}$ & Raw dataset from CSV file with point-based time series. \\
$T_{\text{in}}$ & Length of the input time series window (300 steps). \\
$T_{\text{out}}$ & The specific future time step targeted for prediction. \\
$H, W$ & Height and width of the spatial grid (256$\times$256). \\
$\mathbf{p}$ & Set of 2D coordinates (Easting, Northing) for measurement points. \\
$\mathcal{G}$ & The regular $H \times W$ grid used for spatial interpolation. \\

\midrule
\multicolumn{2}{l}{\textbf{Data Structures}} \\
$\mathcal{X}$ & Input data tensor of shape ($T_{\text{in}}, H, W$) after gridding. \\
$\mathbf{X}_t$ & A single dense displacement map at an input time step $t$. \\
$\mathbf{Y}_{\text{true}}$ & The ground truth displacement map for the target time $T_{\text{out}}$. \\
$\mathcal{X}_{\text{5D}}$ & 5D input tensor for the CNN-LSTM model. \\
$\mathbf{X}_{\text{table}}$ & Tabular version of $\mathcal{X}$, with pixels as rows and time steps as columns. \\
$\mathbf{y}_{\text{true}}$ & Vectorized version of the ground truth map $\mathbf{Y}_{\text{true}}$. \\

\midrule
\multicolumn{2}{l}{\textbf{Models \& Outputs}} \\
$\mathcal{M}$ & The set of all models: $\{\mathcal{M}_{\text{CNN-LSTM}}, \mathcal{M}_{\text{LGBM}}, \mathcal{M}_{\text{Lasso}}\}$. \\
$\mathcal{M}_{\text{CNN-LSTM}}$ & The trained CNN-LSTM model. \\
$\mathcal{M}_{\text{LGBM}}$ & The trained LightGBM model. \\
$\mathcal{M}_{\text{Lasso}}$ & The trained Lasso regression model. \\
$\hat{\mathbf{Y}}_m$ & The predicted displacement map generated by a model $m \in \mathcal{M}$. \\

\midrule
\multicolumn{2}{l}{\textbf{Evaluation \& Interpretation}} \\
$\text{RMSE}_m$ & Root Mean Squared Error for a model $m$. \\
$\text{R}^2_m$ & Coefficient of determination (R-squared) for a model $m$. \\
$\mathcal{E}$ & The SHAP TreeExplainer object for the LightGBM model. \\
$\mathbf{\Phi}_{\text{LGBM}}$ & Matrix of SHAP values calculated for the LightGBM model. \\
\bottomrule
\end{tabular}
\caption{Notation Key for the Forecasting and Analysis Pipeline.}
\label{tab:main_notation}
\end{table}

\subsection{Data generation}
\label{DataGeneration}

Our methodology for spatio-temporal prediction of ground displacement begins with a significant data preprocessing stage. The raw EGMS dataset provides displacement values for a set of sparsely distributed Measurement Points (MPs), each identified by its 'northing' and 'easting' coordinates. For each MP, the dataset includes a time series of 300 displacement measurements recorded between 2018 and 2022.

To prepare these data for a deep learning model capable of capturing both spatial and temporal patterns, we transform the sparse point data into a dense, gridded, spatio-temporal format. This process involves creating a sequence of 'displacement maps'. For each of the 300 time steps, we generate a 2D displacement map by interpolating the displacement values from the irregularly spaced MPs onto a regular 256x256 grid. We employ linear interpolation for this task to estimate the displacement at each grid cell. Any resulting Not-a-Number (NaN) values, which may occur at grid points outside the convex hull of the original MPs, are replaced with zero.

\begin{figure}[htbp]
  \centerline{\includegraphics[scale=0.2]{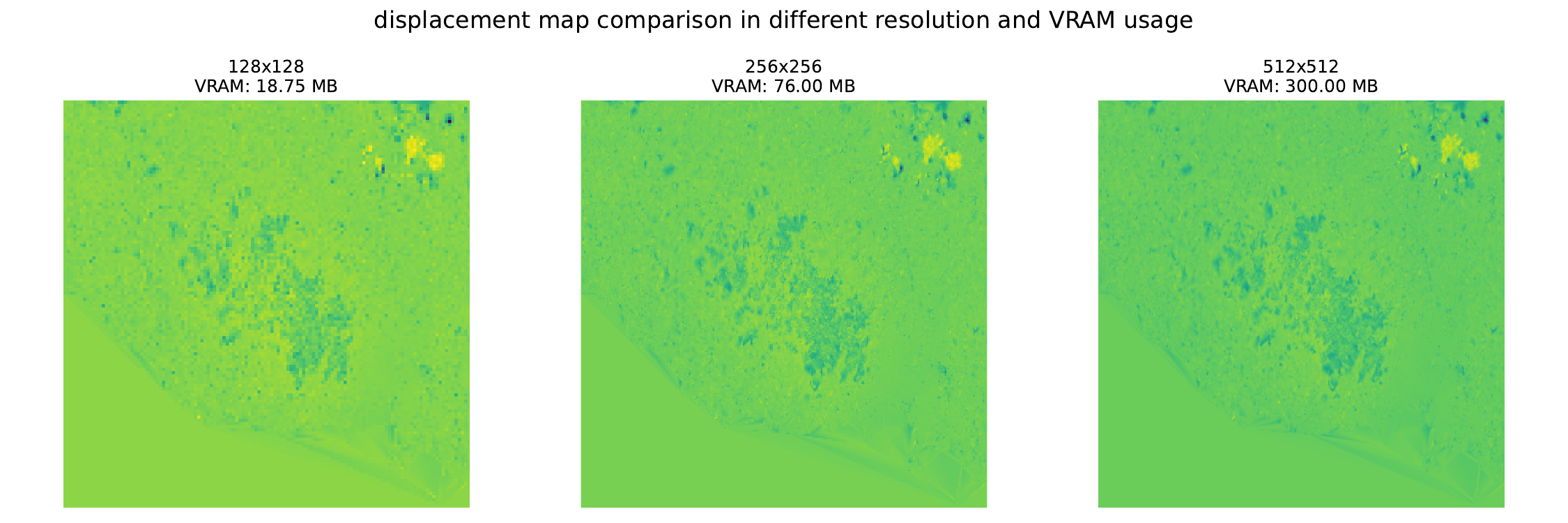}}
  \caption{Visual and computational comparison of different spatial resolutions. Displacement maps are shown for grid sizes of (a) 128x128, (b) 256x256, and (c) 512x512. The VRAM indicates the estimated memory required to load a full training tensor at each resolution, demonstrating the trade-off between image detail and computational demand.}
  \label{fig:VRAM}
  \end{figure}

The choice of the spatial resolution for the interpolated displacement maps is a critical parameter that balances the fidelity of spatial features against computational resource requirements. To determine an optimal resolution, we conducted a comparative analysis of three candidate grid sizes: 128x128, 256x256, and 512x512. The results, including a visual comparison and the corresponding GPU memory (VRAM) footprint for a full input tensor ([1, 300, 1, H, W]), are presented in Figure.~\ref{fig:VRAM}.

Our analysis revealed the following:
\begin{enumerate}
    \item At 128x128 resolution, there is a noticeable loss of fine-grained spatial detail, with key deformation features appearing pixelated and less distinct. While its VRAM requirement of 18.75 MB is minimal, the sacrifice in data quality was deemed too significant.
    \item At 512x512 resolution, while there is a marginal increase in visual sharpness compared to 256x256, the computational cost increases disproportionately. The VRAM footprint rises four-fold to 300.00 MB. This substantial memory demand would severely limit model complexity and batch size, making extensive training and hyperparameter tuning impractical.
    \item The 256x256 resolution emerges as the optimal compromise. As shown in the figure, it captures the essential spatial patterns of ground deformation with high fidelity, showing clear improvements over the 128x128 grid. Crucially, it achieves this quality with a manageable VRAM footprint of 76.00 MB.
\end{enumerate}

Therefore, we selected the 256x256 resolution for all subsequent experiments. This choice ensures that our model is trained on data that is both rich in spatial detail and computationally feasible, enabling efficient and stable training.

The result of this procedure is a stack of 300 sequential displacement maps that form a single data sample with dimensions corresponding to time, height, and width. This spatio-temporal tensor allows our model not only to analyze the time series evolution of displacement at each location but also to learn from the spatial relationships between adjacent grid cells at any given time step. The target displacement map for prediction is generated using the same interpolation process as used for the final recorded displacement values.
\subsection{Framework for Single-Step-Ahead Forecasting}
Our study is designed as a single-step-ahead forecasting task, where the model learns from a sequence of 300 historical displacement maps to predict the map at the single, subsequent time step. This specific framework was intentionally chosen for several key reasons.

Firstly, it establishes a clear and rigorous benchmark for evaluating the core spatio-temporal learning capabilities of the proposed CNN-LSTM architecture against the baseline models. By focusing on the most immediate future state, we can directly assess the model's ability to extrapolate from the learned historical dynamics without introducing the additional complexities of long-term forecasting, such as the accumulation of errors.

Secondly, this approach allows us to isolate the primary variable of interest in this study: the effectiveness of different model architectures in capturing spatio-temporal dependencies. While multi-step forecasting is an important avenue for future research, a single-step framework provides a more controlled environment to validate the fundamental premise of our paper—that an integrated spatio-temporal model is superior to models that do not explicitly handle both domains. Therefore, predicting the final time step serves as the foundational test of our proposed method's capabilities.
\subsection{Machine Learning Models for Deformation Forecasting}
\label{MLIntro}
To predict future ground displacement, we employ and compare three different machine learning models. Each model approaches the spatio-temporal forecasting problem with a different architectural philosophy. We propose a bespoke deep learning model, CNN-LSTM, which is designed to learn from both spatial and temporal patterns simultaneously. For comparison, we benchmark its performance against two powerful and widely-used regression models: LightGBM and Lasso Regression. These models treat the problem as a high-dimensional regression task, where each pixel time series is an independent sample. The specifics of each model are detailed below.

\begin{enumerate}
    \item \textbf{CNN-LSTM}: The Convolutional Neural Network Long-Short Term Memory (CNN-LSTM) model is a deep learning architecture specifically designed to capture intertwined spatial and temporal dependencies. Our implementation processes the input data, structured as a sequence of $T_{\text{in}}$ displacement maps (a 5D tensor $\mathcal{X}_{\text{5D}}$ of shape $[1, T_{\text{in}}, 1, H, W]$), through two main stages:
    
    \begin{itemize}
        \item \textbf{Spatial Feature Extraction (CNN)}: The CNN component acts as a spatial feature extractor. For each time step in the input sequence, a series of convolutional and max-pooling layers are applied to the $H \times W$ displacement map. Our network consists of three convolutional blocks that progressively reduce the spatial dimensions from $256 \times 256$ down to $32 \times 32$ while increasing the feature depth (from 1 to 128 channels). This hierarchical process allows the model to learn spatial features ranging from simple local textures to more complex regional patterns of deformation.
        
        \item \textbf{Temporal Feature Extraction (LSTM)}: The sequence of flattened feature vectors extracted by the CNN is then fed into an LSTM layer. The LSTM is a specialized Recurrent Neural Network (RNN) architecture designed to learn long-range dependencies in sequential data \cite{LSTMorigin}. It utilizes input, output, and forget gates to regulate the flow of information through time, enabling it to effectively model the temporal evolution of the spatial features. We use the hidden state from the final time step of the LSTM sequence as a comprehensive representation of the entire input history.
    \end{itemize}
    
Finally, a fully-connected linear layer maps this final hidden state to the desired output shape, producing a complete predicted displacement map $\hat{\mathbf{Y}}_{\text{CNN-LSTM}}$ of size $H \times W$. The entire model is trained end-to-end by minimizing the Mean Squared Error (MSE) between the predicted and target displacement maps using the Adam optimizer.

    \item \textbf{LightGBM}: LightGBM is a highly efficient, gradient-boosting decision tree (GBDT) framework \cite{LightGBM}. Unlike the CNN-LSTM, LightGBM does not inherently process spatial data. Instead, we transform the input data into a tabular format, $\mathbf{X}_{\text{table}}$, where each row corresponds to a single pixel and the columns represent the displacement values over the $T_{\text{in}}$ time steps. The model's objective is to predict the target displacement for each pixel independently. It builds an ensemble of decision trees sequentially, where each new tree corrects the errors of the previous ones. We train the model using a portion of the pixels as a training set and evaluate on a hold-out validation set, employing early stopping to prevent overfitting and identify the optimal number of boosting rounds.

    \item \textbf{Lasso Regression}: As a robust linear baseline, we use Lasso (Least Absolute Shrinkage and Selection Operator) Regression \cite{Lasso}. Similar to LightGBM, this model operates on the tabular data matrix $\mathbf{X}_{\text{table}}$. Lasso is a linear regression model that incorporates an L1 regularization term into its loss function. The loss function is defined as the sum of the MSE and the L1-norm of the coefficient vector, weighted by a regularization parameter $\alpha$. This regularization has the effect of shrinking some feature coefficients to exactly zero, effectively performing automatic feature selection. In our context, this helps identify which past time steps are most influential for predicting the future displacement. Our implementation uses PyTorch to train the model on a GPU, optimizing the Lasso loss function with the Adam optimizer.
\end{enumerate}

\subsection{Performance comparison for each model}
\label{PerformanceComparison}
\subsubsection{Usage for comparison metrics}
We used the following parameters to test the model performance when performing the estimation task. Equations and definitions are as follows\cite{MSE}:

\begin{enumerate}
  \item $\textbf{RMSE}(y, \hat{y}) = \sqrt{\frac{\sum_{i=0}^{N - 1} (y_i - \hat{y}_i)^2}{N}}$

  Root Mean Square Error (RMSE) is the square root of the mean squared differences between the predicted and actual values. Lower RMSE indicates better prediction accuracy. 
  \item $\textbf{MSE}(y, \hat{y}) = \frac{\sum_{i=0}^{N - 1} (y_i - \hat{y}_i)^2}{N}$

  Mean Squared Error (MSE) measures prediction error by calculating the average squared difference between the observed and predicted values. A lower MSE means that the model has less error and fits the data better.

  \item $\textbf{R$^2$} = 1-\frac{\sum({y_i}-\hat{y_i})^2}{\sum(y_i-\bar{y})^2}$

  The coefficient of determination (R$^2$) indicates how well the model fits the data. It is essential to evaluate the accuracy of the model in predicting outcomes or testing hypotheses. As a percentage, the value is always between 0 and 1. A higher R$^2$ means more output values are accounted for by the input values.
\end{enumerate}

In our training and testing process, we trained the models on the training set and tested their performance using the above parameters generated by the test set. This helps us to evaluate the performance of the model and the accuracy of the prediction.

\subsubsection{Quantitative and Statistical Error Analysis}
For a more granular and quantitative evaluation, a comprehensive error and residual analysis was performed. This involved generating a suite of diagnostic graphs for each model, as presented in Figure.~\ref{fig:combined_error_analysis}, to systematically dissect the nature and distribution of the prediction errors. The following four metrics were used:

\begin{enumerate}
    \item \textbf{Scatter Plots:} These plots of predicted versus actual displacement values were used to assess the overall correlation and identify any systematic deviations from the ideal one-to-one relationship.
    
    \item \textbf{Residual Plots:} To detect model bias, the residuals (the difference between predicted and true values) were plotted against the true displacement values. This helps reveal whether the model systematically overestimates or underestimates displacement in certain ranges.
    
    \item \textbf{Binned Mean Absolute Error (MAE):} To understand how the performance of the model varies with the magnitude of displacement, the MAE was calculated and plotted for different bins (ranges) of the true displacement values.
    
    \item \textbf{Binned Residuals Boxplots:} To analyze the statistical distribution of the errors, box plots of residuals were generated for each bin of true displacement. This provides a clear view of the median error, the interquartile range (error variance), and the presence of outliers throughout the data range.
\end{enumerate}

Together, these diagnostic plots facilitate a rigorous, multifaceted comparison of the predictive accuracy, bias, and stability of the LASSO, LightGBM, and CNN-LSTM models.
\subsubsection{Visual Assessment of Spatial Prediction}
A qualitative visual analysis was conducted to evaluate each model's ability to reproduce the spatial characteristics of the ground-truth displacement. The predicted displacement maps for the final time step from the LASSO, LightGBM and proposed CNN-LSTM models were compared side by side with the ground-truth map, as shown in Figure .~\ref{fig:sp-perfect-comparison}. This direct comparison serves to assess how effectively each model captures key features, including the shape and boundaries of the primary deformation zone, the internal texture of the displacement field, and any localized anomalies. The primary goal of this assessment is to evaluate the spatial coherence and physical realism of the predictions beyond simple quantitative metrics.

\subsubsection{Usage for correlation analysis for different parameters (SHAP analysis)}
\label{SHAPIntro}

To interpret the temporal dependencies learned by our LightGBM displacement prediction model, we employ SHapley additive exPlanations (SHAP) values - a game-theoretic approach that quantifies feature contributions to model predictions based on Shapley values \cite{lundberg2017unified}. The Shapley value provides a theoretically grounded mechanism to fairly attribute predictive outcomes to individual features by evaluating their marginal contributions across all possible feature coalitions.

\begin{equation}
\phi_i(f,x) = \sum_{S \subseteq F \setminus \{i\}} \frac{|S|! \cdot (|F| - |S| - 1)!}{|F|!} \left[ f_x(S \cup \{i\}) - f_x(S) \right]
\label{eq:shapley}
\end{equation}

where $\phi_i$ is the Shapley value for the feature $i$, $F$ is the complete set of features, $S$ represents a subset of features and $f_{x}(S)$ denotes the prediction of the model when only the features in $S$ are retained (others masked with baseline values).

Our SHAP analysis pipeline, formalized in Algorithm.~\ref{alg:shap_pipeline}, systematically quantifies the contribution of historical displacement values in each timestep to future land deformation predictions. This dual-scale approach reveals both global temporal patterns and pixel-level dynamics critical for geophysical interpretation. Table.~\ref{tab:shap_notation} describes variables used in the algorithm.

\begin{algorithm}[htbp]
\caption{SHAP Interpretation Pipeline for InSAR Displacement Forecasting}
\label{alg:shap_pipeline}
\begin{algorithmic}[1]
\small
\Require  
$\mathcal{M}_{\text{LGB}}$: Trained LightGBM model (Section.~\ref{MLIntro}) \\
$\mathcal{D}_{\text{test}}$: Test dataset $(X_{\text{test}}, y_{\text{test}})$ \\
$K$: Sample size for SHAP computation (10,000 pixels) 

\Ensure  
Global feature importance profiles \\
Local prediction-decomposition plots

\State \textbf{Initialize Tree Explainer}
\State $\mathcal{E} \leftarrow \text{TreeExplainer}(\mathcal{M}_{\text{LGB}})$ 
    \Comment{Computes SHAP values via path-dependent perturbation \cite{lundberg2017unified}}
\State $\phi_{\text{base}} \leftarrow \mathcal{E}.\text{expected\_value}$ 
    \Comment{Baseline prediction (average model output)}

\State \textbf{Compute SHAP Values}
\State $\mathbf{\Phi} \leftarrow \mathcal{E}.\text{shap\_values}(X_{\text{test}}[0:K])$

\State \textbf{Global Feature Analysis}
\State Generate summary plot: $\texttt{shap.summary\_plot}(\mathbf{\Phi}, X_{\text{test}}[0:K], \mathcal{F})$
\State where $\mathcal{F} = \{f_{t-\tau} \mid \tau = T_{\text{in}},\dots,1\}$ 
    \Comment{Features $f_{t-\tau}$: displacement at $\tau$ timesteps before prediction}

\State \textbf{Local Prediction Decomposition}
\State Select representative pixel $p$: $\mathbf{x}_p \leftarrow X_{\text{test}}[p],  \phi_p \leftarrow \mathbf{\Phi}[p]$
\State Generate force plot: $\texttt{force\_plot}(\phi_{\text{base}}, \phi_p, \mathbf{x}_p, \mathcal{F})$
    \Comment{Visualizes feature contributions relative to baseline}

\State \textbf{SAR-Specific Baseline Validation (Optional)}
\If{$\texttt{clutter\_aware} = \text{True}$}
\State $\mathcal{L} \leftarrow \sum_{i=1}^K \sum_{j=1}^{T_{\text{in}}} |\phi_{ij}| + \lambda \cdot \text{MMD}^2(\mathcal{X}_{\text{clutter}}, \mathbf{b})$
    
\EndIf
\end{algorithmic}
\end{algorithm}

\begin{table}[htbp] 
  \centering 
  \small
  \begin{tabular}{p{2.2cm}p{4.5cm}} 
    \toprule 
    \textbf{Symbol} & \textbf{Description} \\ 
    \midrule 
    $\mathcal{M}_{\text{LGB}}$ & Trained LightGBM regression model \\ 
    $\mathcal{D}_{\text{test}}$ & Test dataset (input windows and target displacements) \\ 
    $T_{\text{in}}$ & Input window length (29 timesteps) \\ 
    $\mathcal{E}$ & SHAP TreeExplainer object \\ 
    $\phi_{\text{base}}$ & Baseline prediction value (expected model output) \\ 
    $\mathbf{\Phi}$ & Matrix of SHAP values ($\phi_{ij}$ = contribution of timestep $j$ to sample $i$) \\ 
    $\mathcal{F}$ & Feature set: historical displacements $\{f_{t-29}, f_{t-28}, \dots, f_{t-1}\}$ \\ 
    $\mathbf{x}_p$ & Input window for a specific pixel $p$ \\ 
    $\mathcal{X}_{\text{clutter}}$ & Bac kground clutter distribution (e.g., Rayleigh/Weibull) \\ 
    $\text{MMD}$ & Maximum Mean Discrepancy distributional distance metric \\ 
    \bottomrule 
  \end{tabular}
  \caption{Notation Key for SHAP Analysis} 
  \label{tab:shap_notation} 
\end{table}

\section{Results}
\label{sec:Results}
\subsection{Performance assessment using classification metrics}
\label{subsec:classification}
To complement the visual analysis, we computed the Root Mean Squared Error (RMSE), Mean Squared Error (MSE), and the coefficient of determination (R$^2$) for each model, with the results summarized in Table.~\ref{tab:parameter}.

\begin{table}[htbp]{}
        \centering
        \begin{tabular}{cccc}
            \toprule
            models & RMSE & MSE & R$^2$ \\
            \midrule
            LASSO Regression  & 1.2101  &  1.4644 &  0.9504 \\
            CNN-LSTM  &  0.2854 & 0.5342  & 0.9901 \\
            Lightgbm  & 1.3005  & 1.6914 & 0.9424 \\
            \bottomrule
        \end{tabular}
        \caption{RMSE, MSE, and R$^2$ for each model}
        \label{tab:parameter}
    \end{table}
The CNN-LSTM model emerges as the superior architecture, achieving an RMSE of 0.2854 and an R$^2$ value of 0.9901. This indicates an extremely high degree of accuracy and a near-perfect fit to the target data, confirming that its ability to learn from both spatial and temporal features yields the most precise predictions.

The LASSO Regression and LightGBM models, while still achieving high R$^2$ values of 0.9504 and 0.9424 respectively, exhibit significantly higher errors. The LASSO model's RMSE of 1.2101 is more than four times higher than that of the CNN-LSTM, and the LightGBM model performs the poorest with an RMSE of 1.3005. These higher error values align with the visual discrepancies observed in their respective prediction maps, such as magnitude offsets and noisy artifacts.

\subsection{Comparative Error Analysis}
\label{subsec:CEA}
\begin{figure*}[t!]
    \centering
    % Row 1: LASSO Regression
    \begin{subfigure}[b]{0.24\textwidth}
        \includegraphics[width=\textwidth]{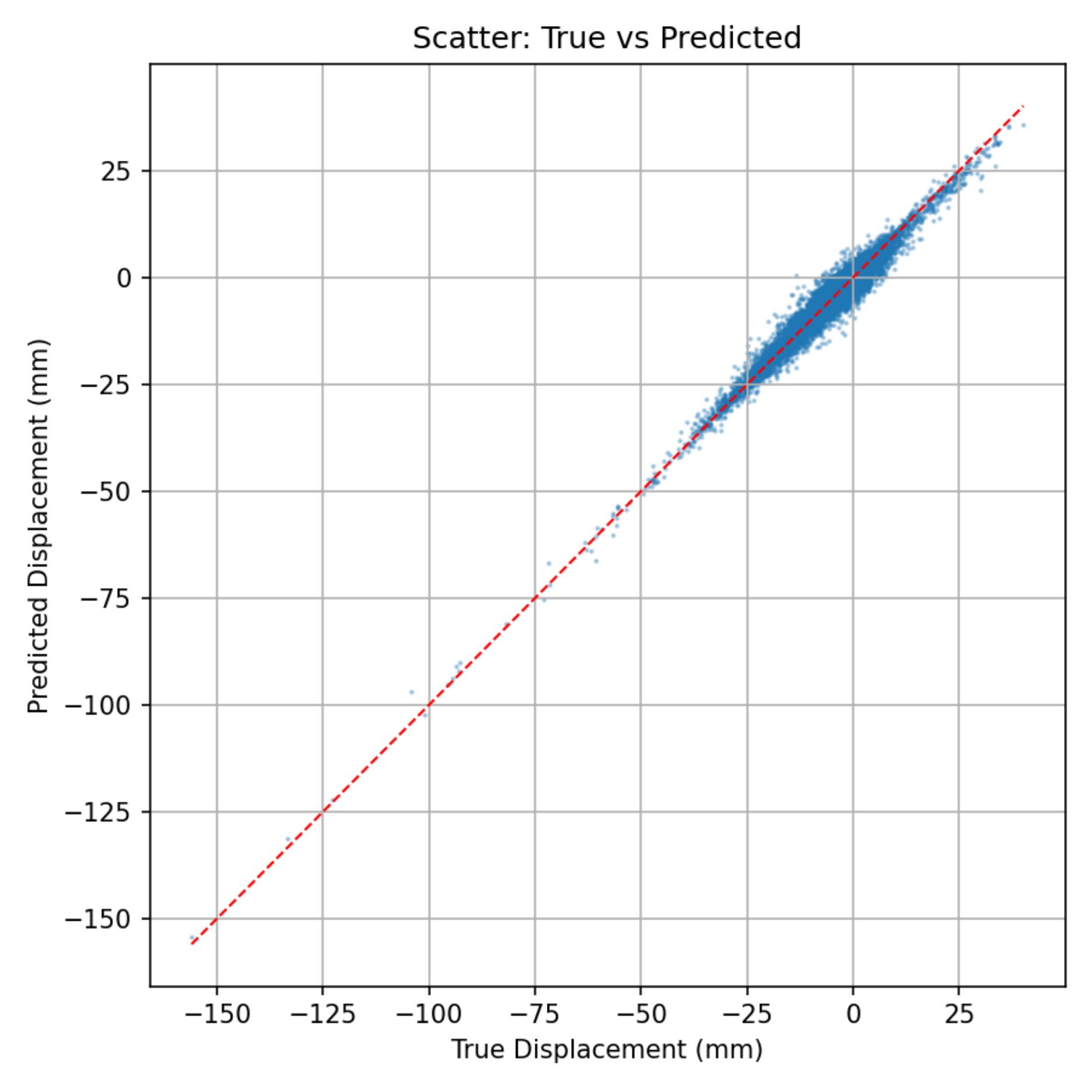}
        \caption{LASSO: Scatter}
        \label{fig:las_scatter_combo}
    \end{subfigure}
    \hfill
    \begin{subfigure}[b]{0.24\textwidth}
        \includegraphics[width=\textwidth]{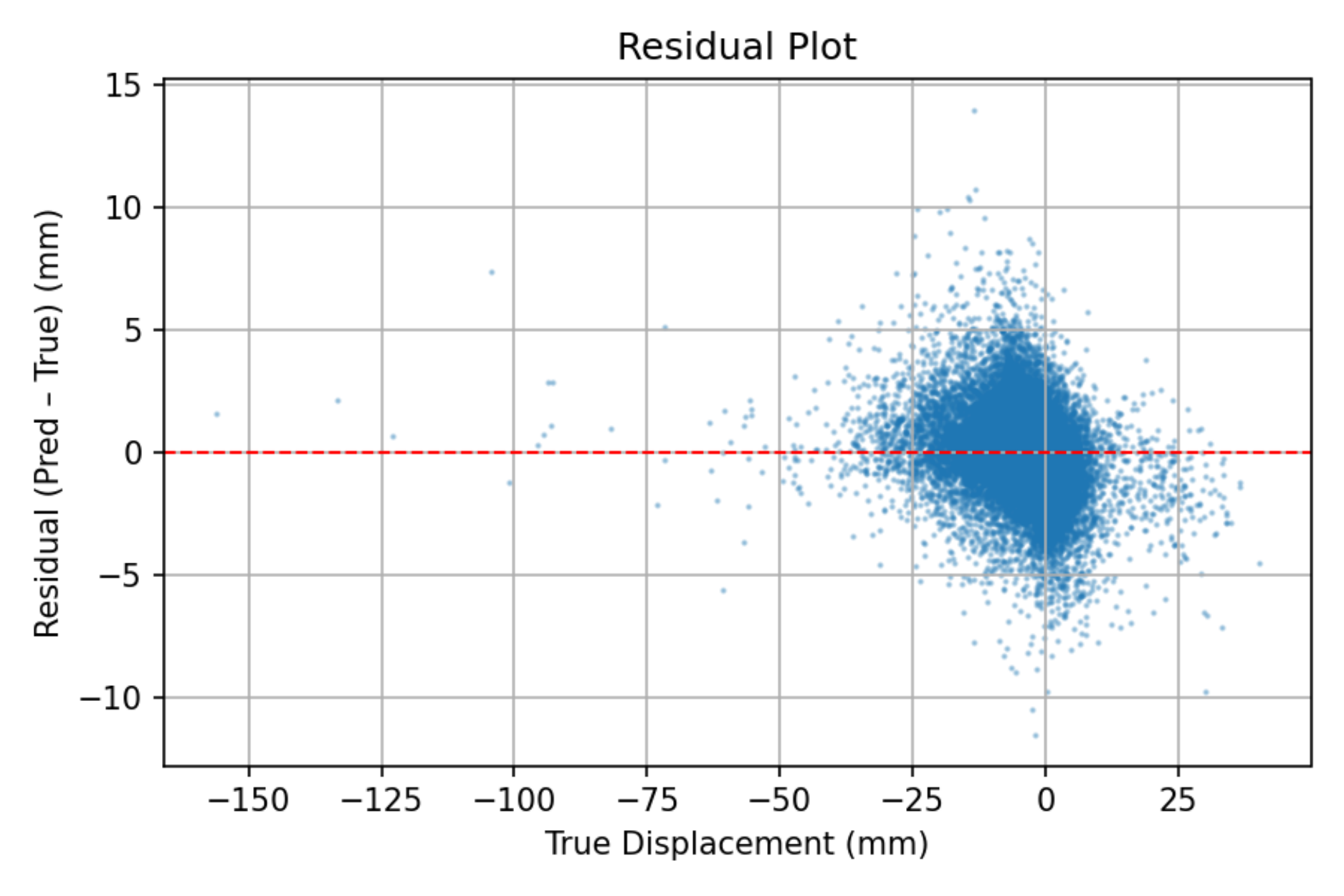}
        \caption{LASSO: Residual}
        \label{fig:las_residual_combo}
    \end{subfigure}
    \hfill
    \begin{subfigure}[b]{0.24\textwidth}
        \includegraphics[width=\textwidth]{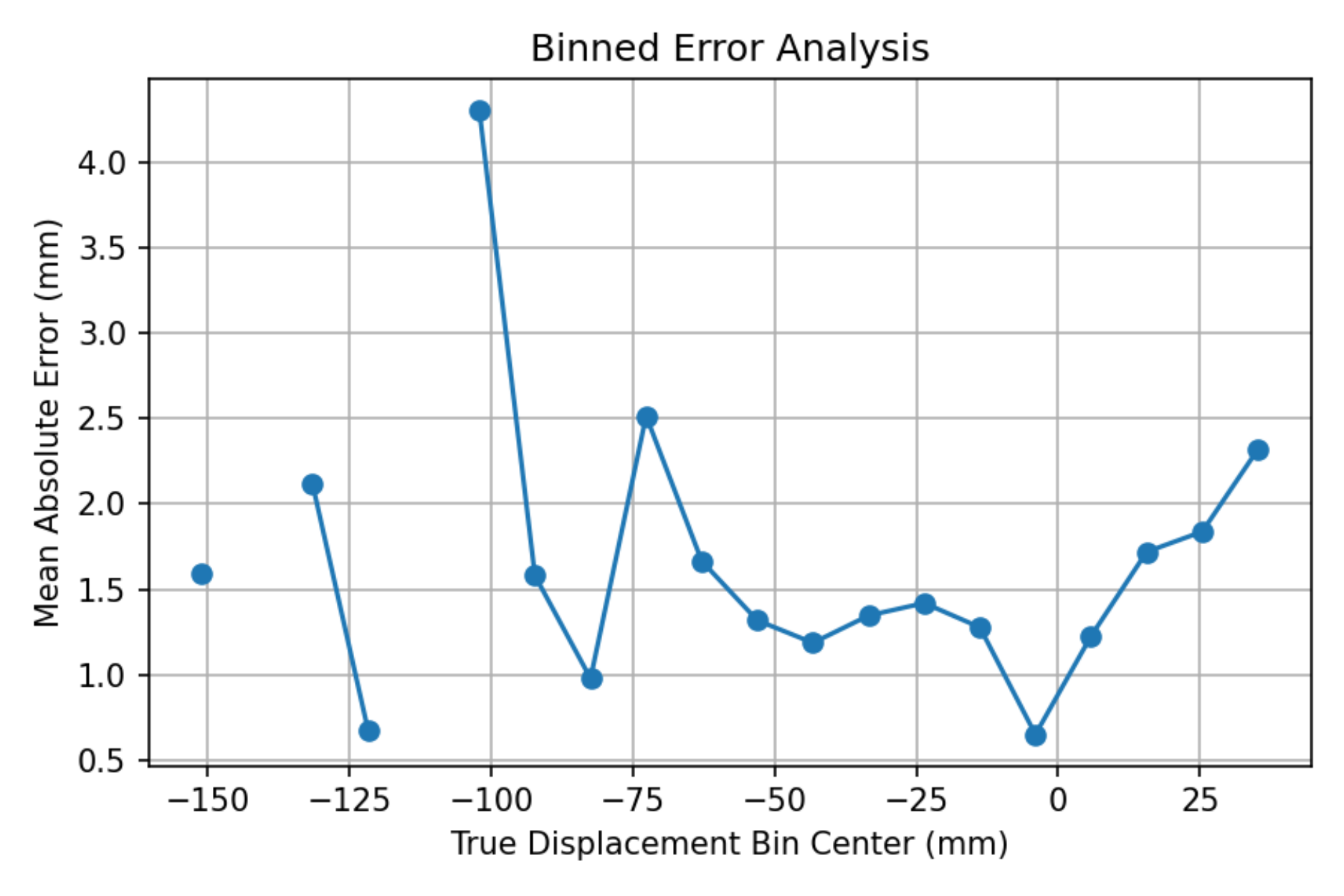}
        \caption{LASSO: Binned Error}
        \label{fig:las_binned_error_combo}
    \end{subfigure}
    \hfill
    \begin{subfigure}[b]{0.24\textwidth}
        \includegraphics[width=\textwidth]{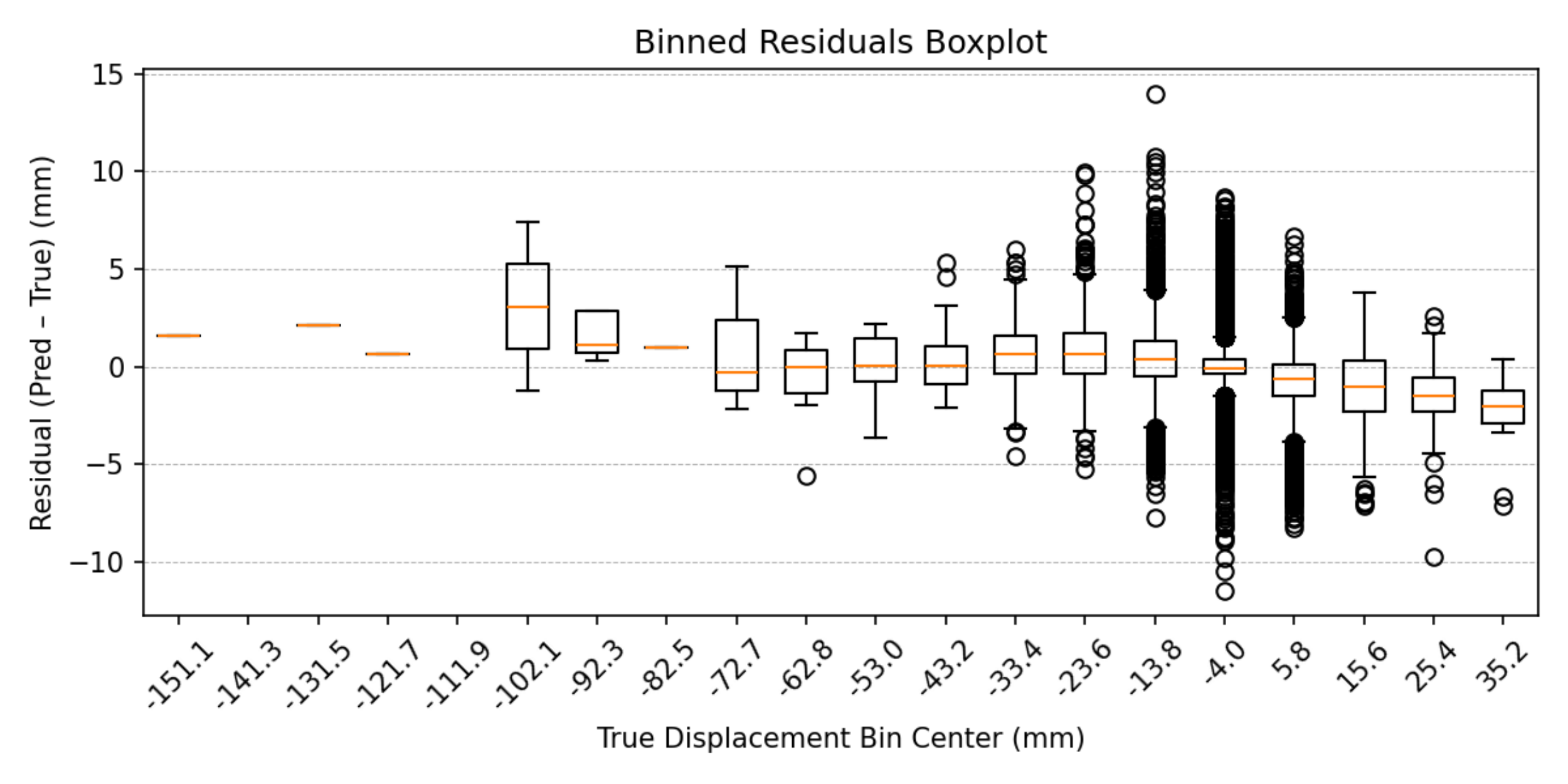}
        \caption{LASSO: Binned Boxplot}
        \label{fig:las_binned_boxplot_combo}
    \end{subfigure}

    \vspace{1em} % Add some vertical space between rows

    % Row 2: LightGBM
    \begin{subfigure}[b]{0.24\textwidth}
        \includegraphics[width=\textwidth]{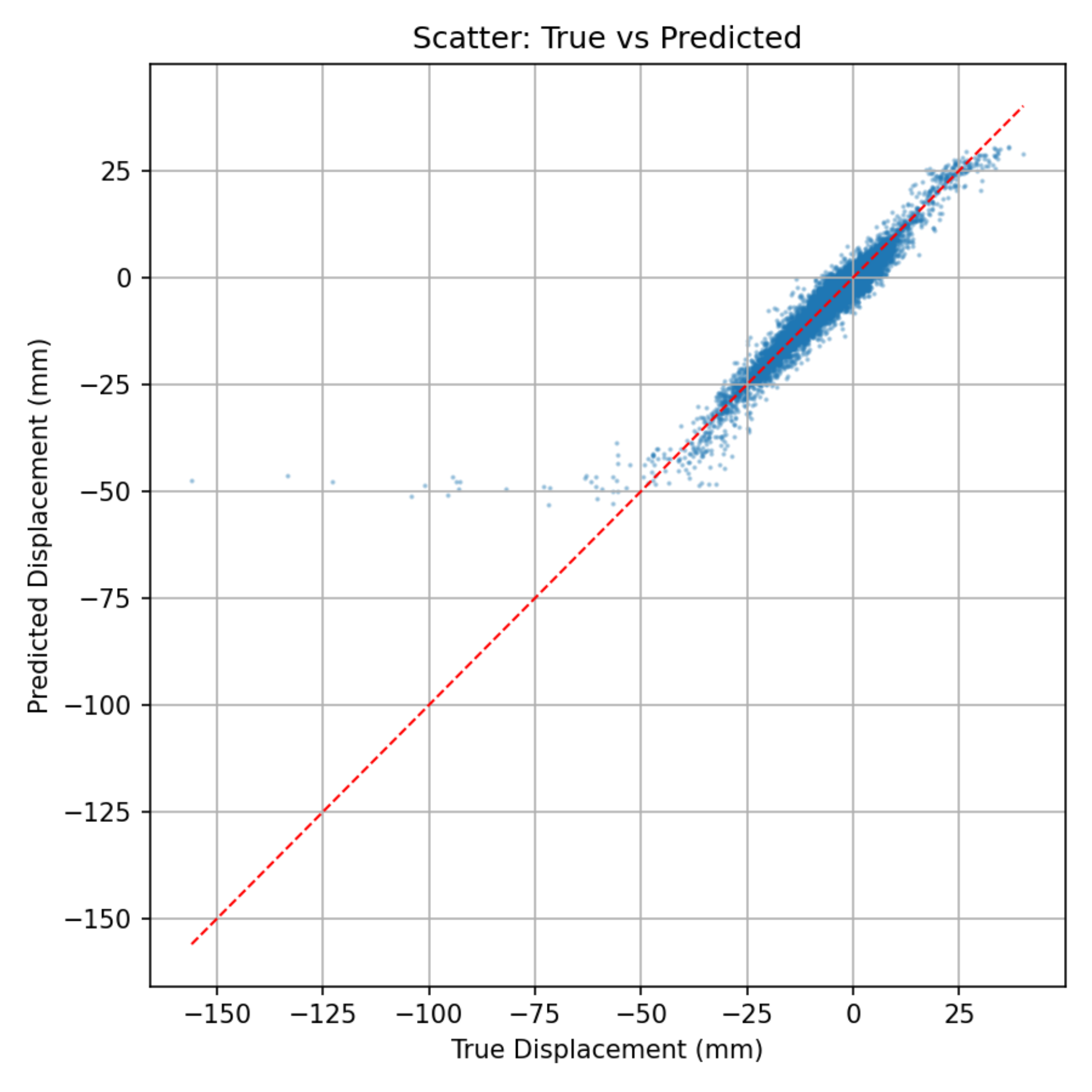}
        \caption{LightGBM: Scatter}
        \label{fig:lgb_scatter_combo}
    \end{subfigure}
    \hfill
    \begin{subfigure}[b]{0.24\textwidth}
        \includegraphics[width=\textwidth]{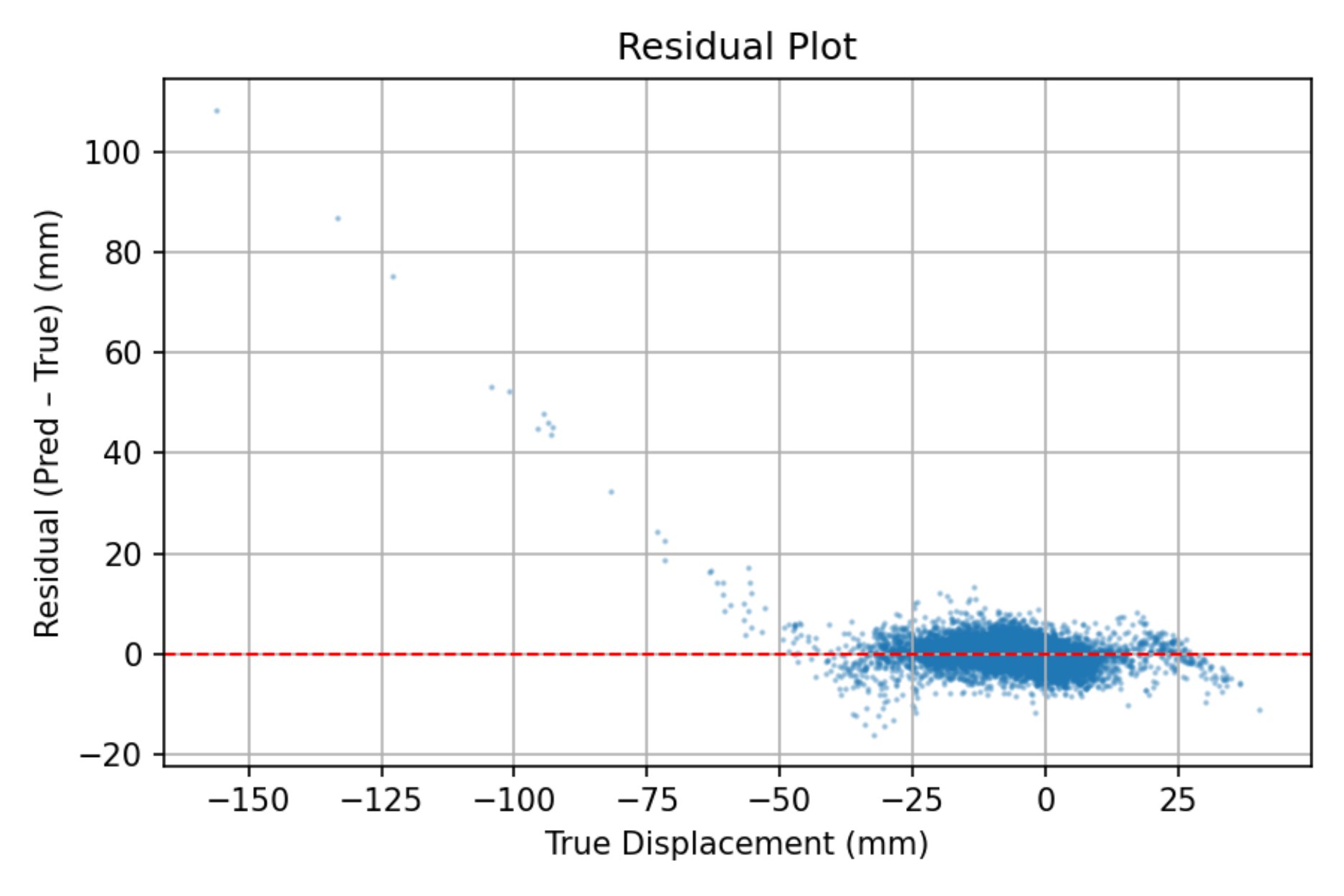}
        \caption{LightGBM: Residual}
        \label{fig:lgb_residual_combo}
    \end{subfigure}
    \hfill
    \begin{subfigure}[b]{0.24\textwidth}
        \includegraphics[width=\textwidth]{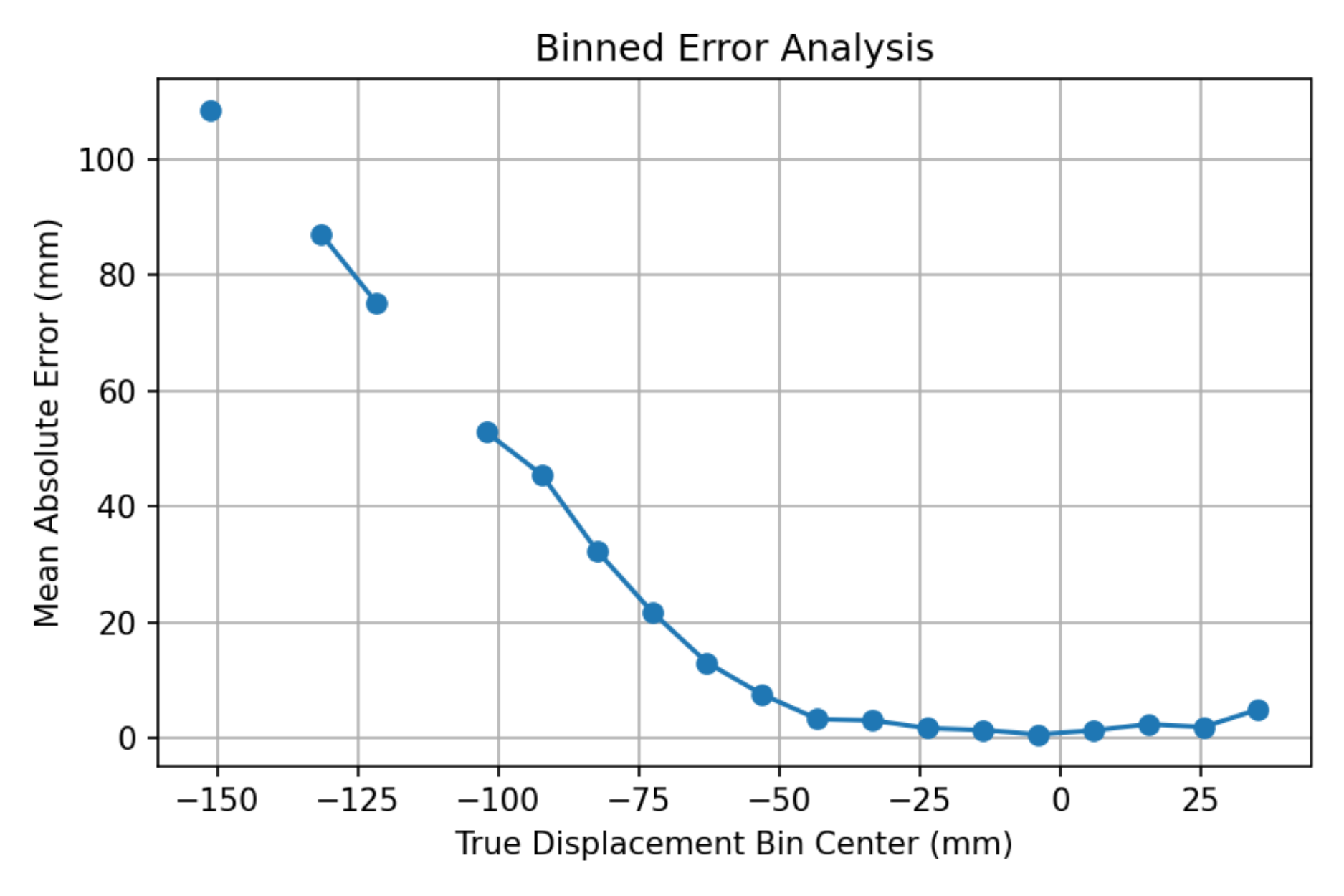}
        \caption{LightGBM: Binned Error}
        \label{fig:lgb_binned_error_combo}
    \end{subfigure}
    \hfill
    \begin{subfigure}[b]{0.24\textwidth}
        \includegraphics[width=\textwidth]{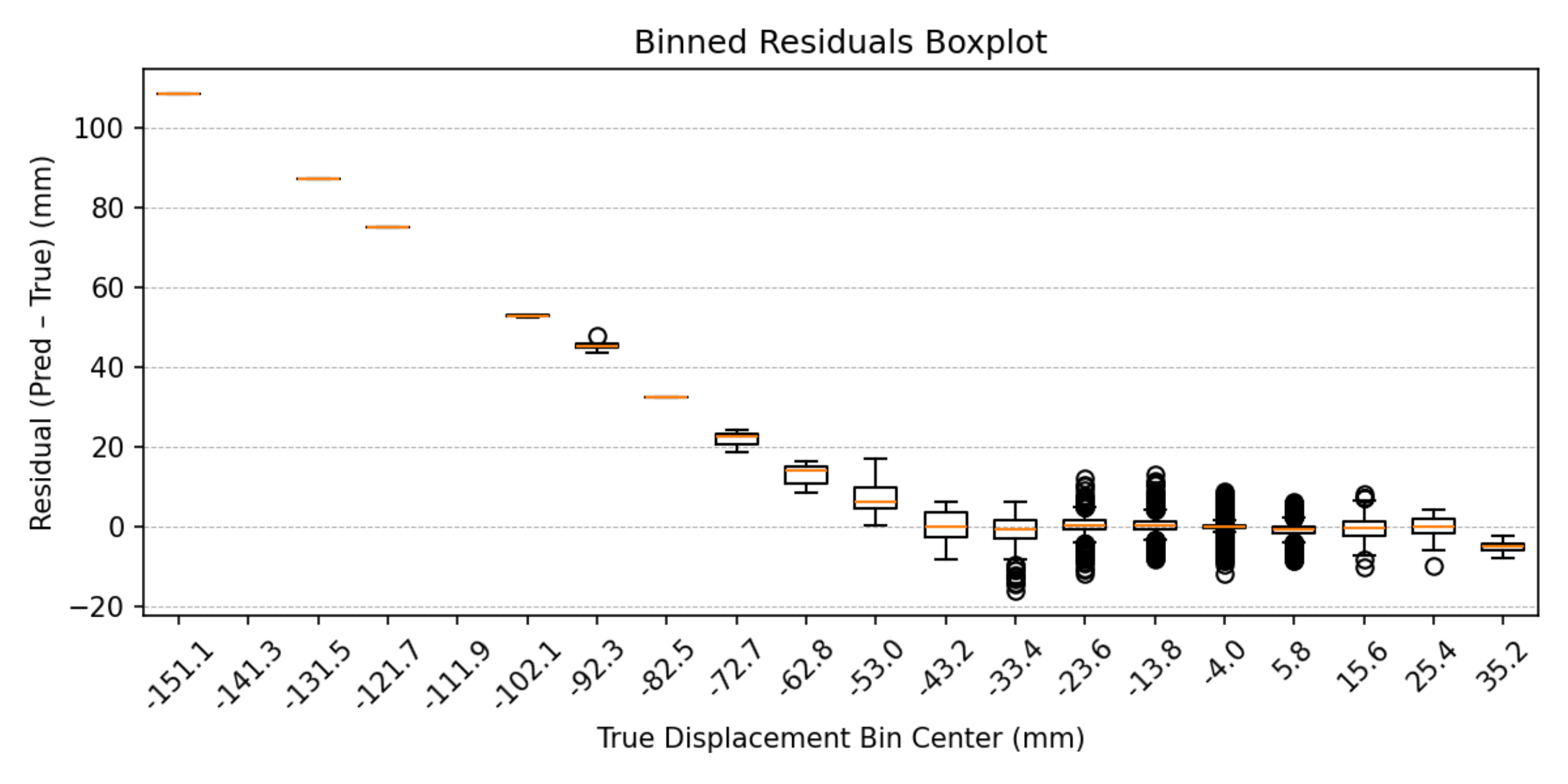}
        \caption{LightGBM: Binned Boxplot}
        \label{fig:lgb_binned_boxplot_combo}
    \end{subfigure}

    \vspace{1em} % Add some vertical space between rows

    % Row 3: CNN-LSTM
    \begin{subfigure}[b]{0.24\textwidth}
        \includegraphics[width=\textwidth]{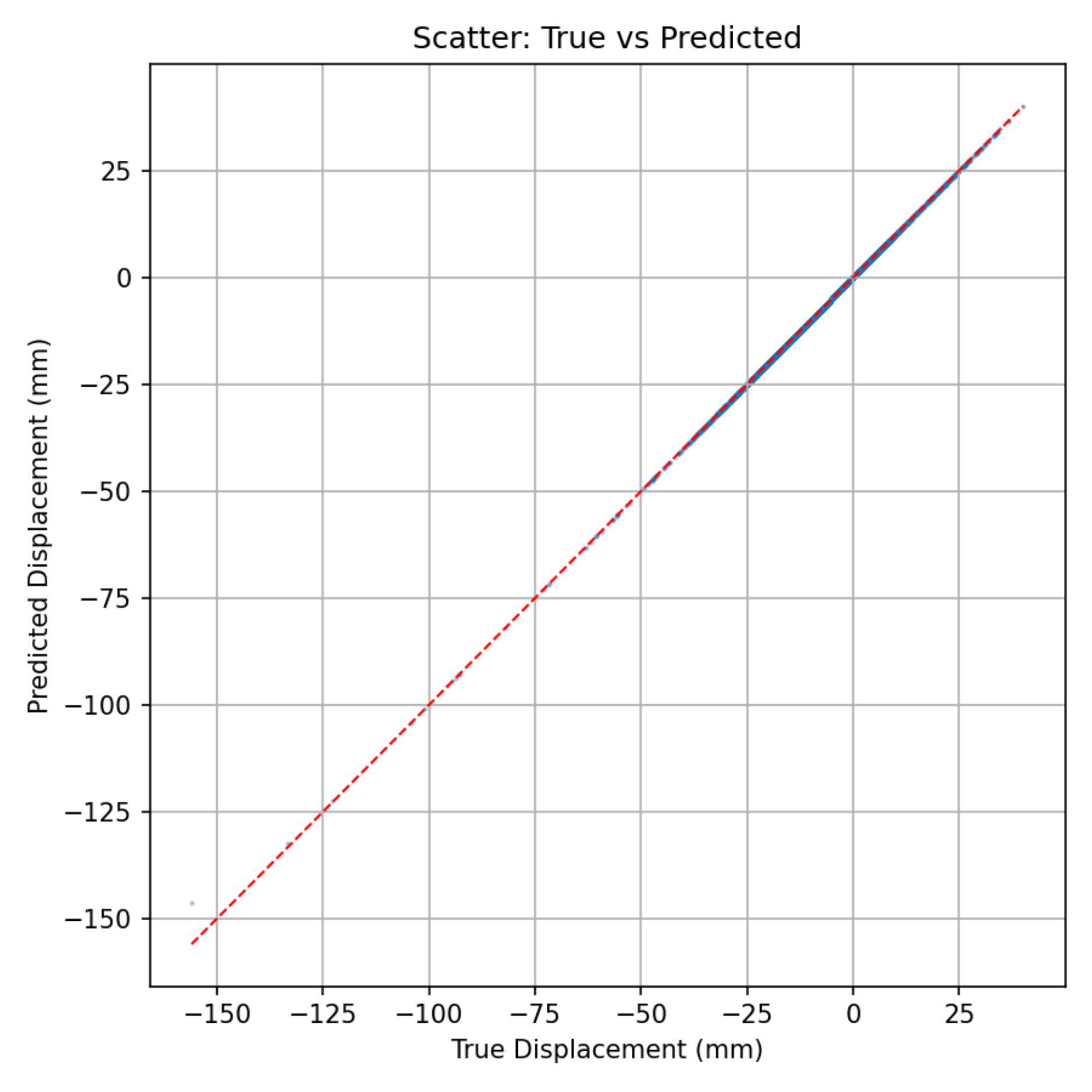}
        \caption{CNN-LSTM: Scatter}
        \label{fig:clstm_scatter_combo}
    \end{subfigure}
    \hfill
    \begin{subfigure}[b]{0.24\textwidth}
        \includegraphics[width=\textwidth]{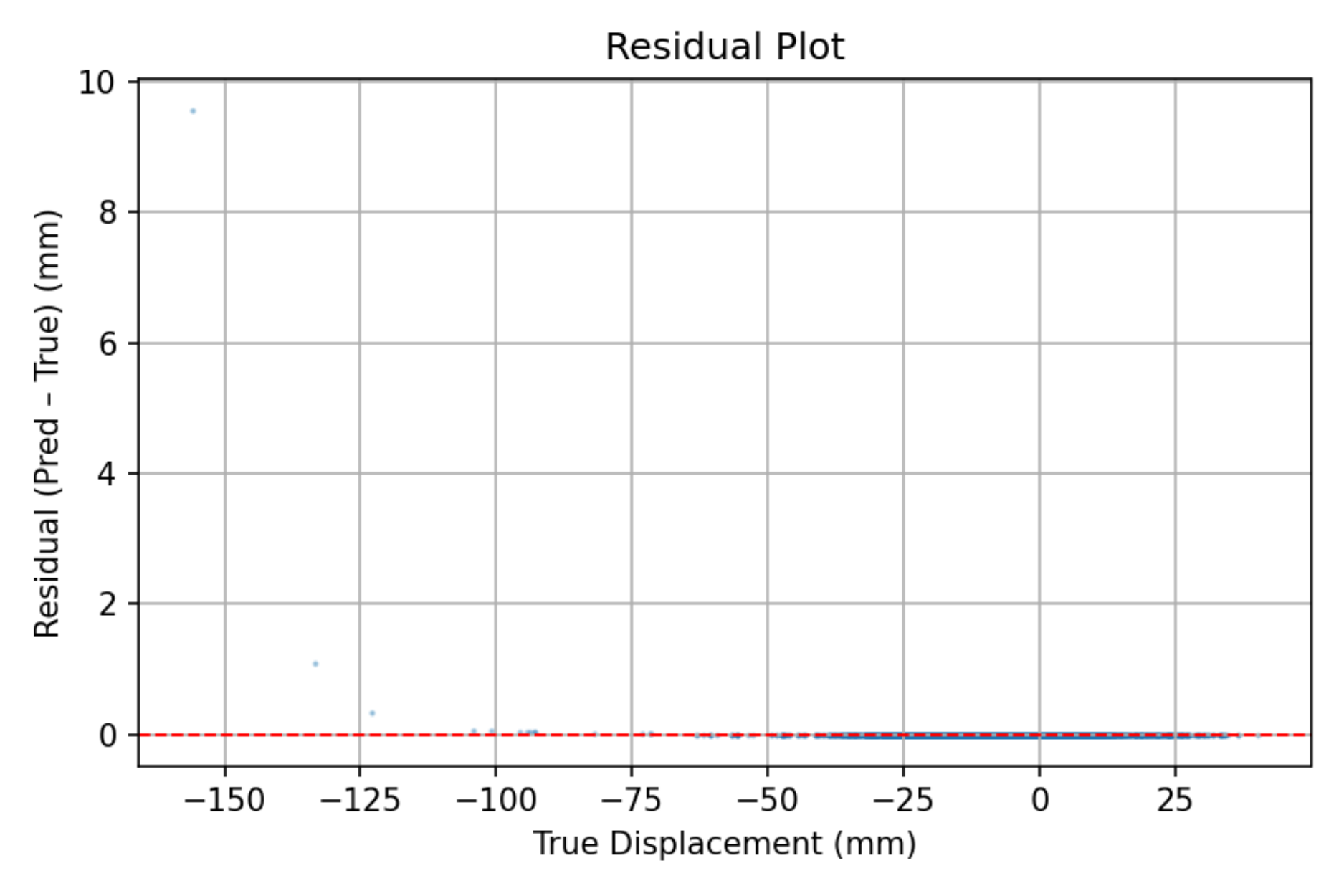}
        \caption{CNN-LSTM: Residual}
        \label{fig:clstm_residual_combo}
    \end{subfigure}
    \hfill
    \begin{subfigure}[b]{0.24\textwidth}
        \includegraphics[width=\textwidth]{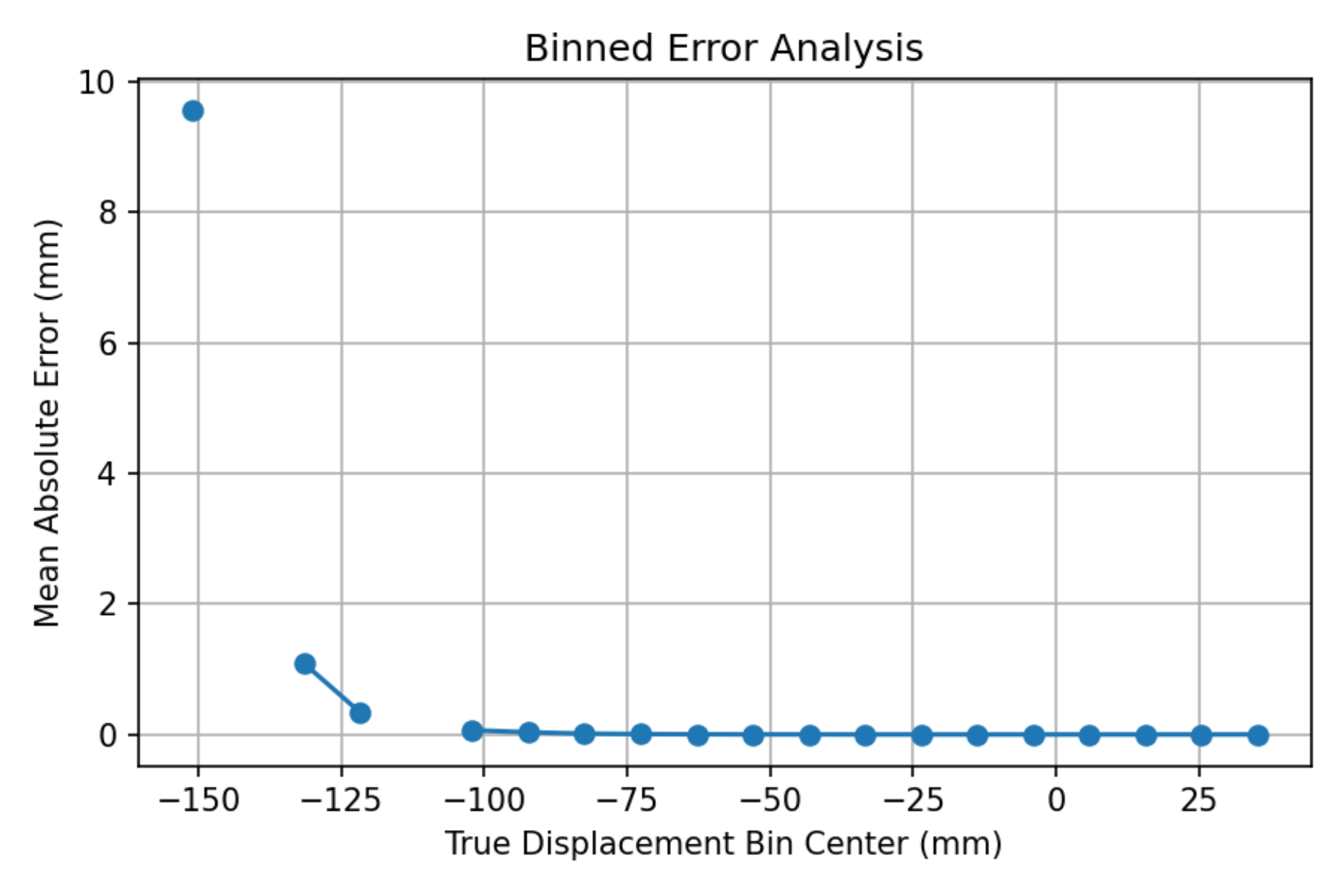}
        \caption{CNN-LSTM: Binned Error}
        \label{fig:clstm_binned_error_combo}
    \end{subfigure}
    \hfill
    \begin{subfigure}[b]{0.24\textwidth}
        \includegraphics[width=\textwidth]{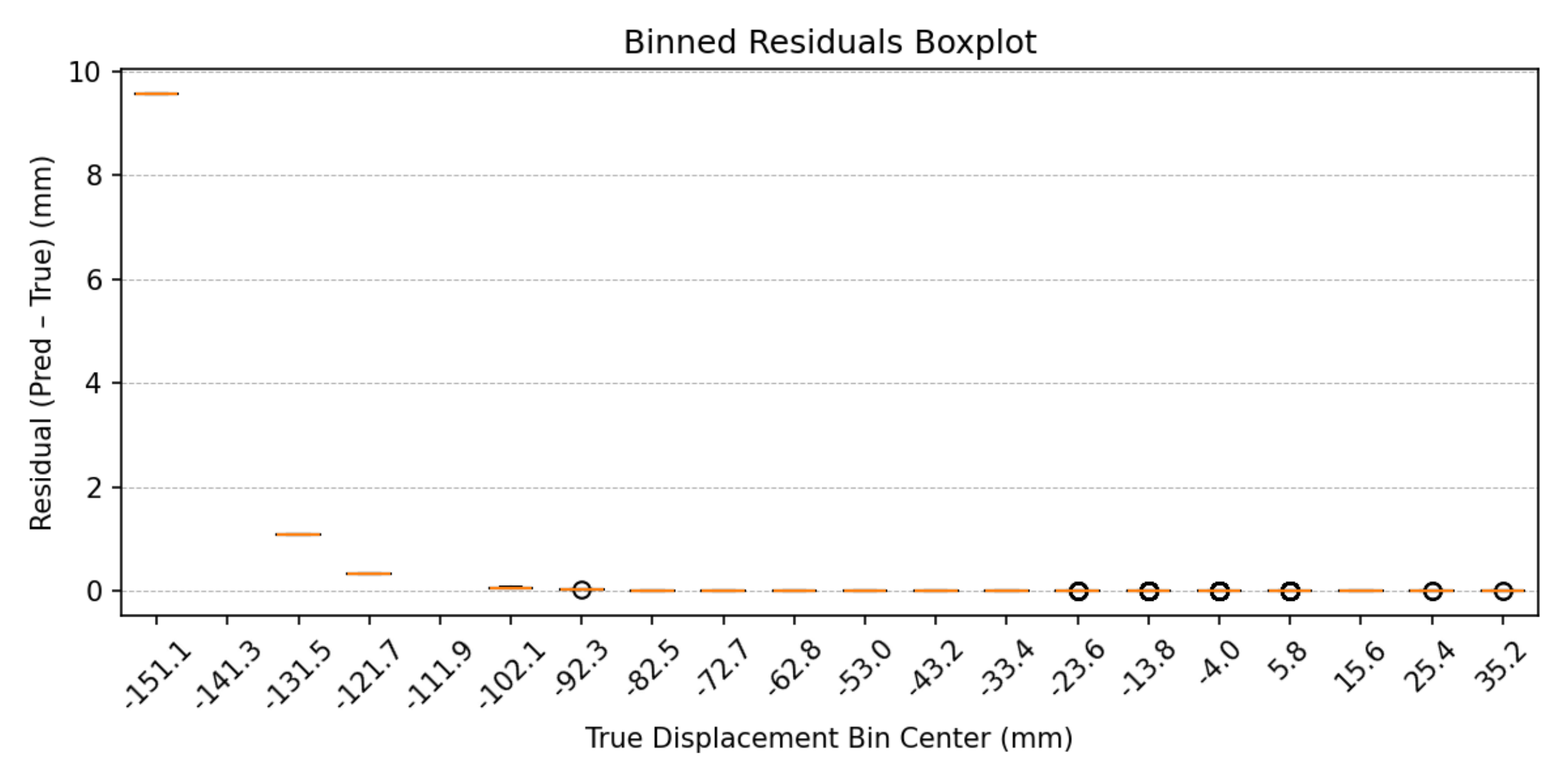}
        \caption{CNN-LSTM: Binned Boxplot}
        \label{fig:clstm_binned_boxplot_combo}
    \end{subfigure}

    \caption{Comprehensive error analysis for all three models. The top row (a-d) displays diagnostics for LASSO Regression. The middle row (e-h) shows diagnostics for LightGBM. The bottom row (i-l) presents the results for the proposed CNN-LSTM model. The columns, from left to right, represent: Scatter Plot (Predicted vs. Actual), Residual Plot, Binned Error Plot, and Binned Residuals Boxplot. The superior performance of the CNN-LSTM model is evident across all metrics, particularly in the tightly clustered scatter plot (i) and the randomly distributed, low-magnitude residuals (j).}
    \label{fig:combined_error_analysis}
\end{figure*}

To perform a granular evaluation of the model performance, a detailed comparative error analysis was performed. Figure.~\ref{fig:combined_error_analysis} provides a comprehensive set of diagnostic plots, structured for direct comparison. We assess performance across each diagnostic metric (columns) to highlight the relative strengths and weaknesses of each model.
\subsubsection*{Column 1: Scatter Plots (Predicted vs. Actual)}
The scatter plots reveal the fundamental correlation between the predicted and actual values. The LASSO model, Figure.~\ref{fig:las_scatter_combo}, establishes a baseline, showing a clear linear trend but with significant spread around the identity line. The performance of the LightGBM model, Figure.~\ref{fig:lgb_scatter_combo}, represents a severe degradation, with a much wider dispersion and a clear curvilinear pattern, indicating systematic errors. In contrast, the CNN-LSTM plot, Figure .~\ref{fig:clstm_scatter_combo}, is exemplary, with data points forming a tight, narrow line, visually confirming its superior predictive accuracy.

\subsubsection*{Column 2: Residual Plots}
The residual plots expose the underlying biases in the models. It reveals the residual value (difference between the prediction and the real displacement value) according to the variation of the true displacement value. The LASSO residuals , shown in Figure.~\ref{fig:las_residual_combo} are clustered around zero but in a non-uniform shape, suggesting inconsistent error variance. The LightGBM plot, Figure.~ \ref{fig:lgb_residual_combo}, displays a clear, systematic curvilinear bias. This pattern shows that the model consistently underestimates the magnitude of large negative displacements, a critical failure for this application. The CNN-LSTM model's residual plot, Figure.~\ref{fig:clstm_residual_combo}, however, shows a tight, random, and unbiased distribution of errors along the zero line, indicating a robust and reliable model.

\subsubsection*{Column 3: Binned Mean Absolute Error}
This analysis quantifies the error across different ranges of true displacement. It reveals the Mean Absolute Value (MAE) according to the variation of the true displacement value. Although the LASSO model error plot, Figure.~\ref{fig:las_binned_error_combo}, is relatively low, it fluctuates, peaking at the extremes. The LightGBM plot , Figure.~\ref{fig:lgb_binned_error_combo}, quantifies its failure to predict large subsidence, with the mean absolute error skyrocketing to over 100~mm. The CNN-LSTM plot, Figure.~\ref{fig:clstm_binned_error_combo}, again demonstrates its superiority, with the error remaining close to zero in nearly the entire range of data, proving its effectiveness and stability.

\subsubsection*{Column 4: Binned Residuals Boxplots}
The binned residual boxplots provide a distributional view of the model errors across different ranges of true displacement. This analysis is particularly useful for identifying systematic biases and variations in error magnitude.

The LASSO plot, Figure.~\ref{fig:las_binned_boxplot_combo} shows moderate error variance and reveals a significant number of outliers in several bins, suggesting inconsistent performance. In contrast, the LightGBM boxplot, Figure.~\ref{fig:las_binned_boxplot_combo}, visually confirms the model's systemic bias; for large negative displacements, the boxes are large and heavily skewed upward, indicating that the model consistently and significantly underestimates subsidence.

Finally, the CNN-LSTM plot, Figure.~\ref{fig:clstm_binned_boxplot_combo}, demonstrates exceptional performance. Its boxplots are compressed into flat lines near zero across nearly all displacement bins. This indicates negligible error variance and a lack of significant outliers, confirming the model is robust, unbiased, and highly reliable across the entire range of observed ground movement.

\subsubsection*{Summary of Analysis}
This column-by-column comparison consistently demonstrates the superiority of the proposed CNN-LSTM model. Across every diagnostic metric, from overall correlation and bias detection to binned error analysis, CNN-LSTM outperforms baseline models, which suffer from moderate inaccuracies (LASSO) or severe systematic failures (LightGBM).
\subsection{Visual Comparison of Predicted and Ground Truth Displacement Maps}
\label{subsec:Visual}
\begin{figure*}[t!]
\centering

% Subfigure for LASSO
\begin{subfigure}{\textwidth}
    \centering
    \includegraphics[width=0.9\textwidth]{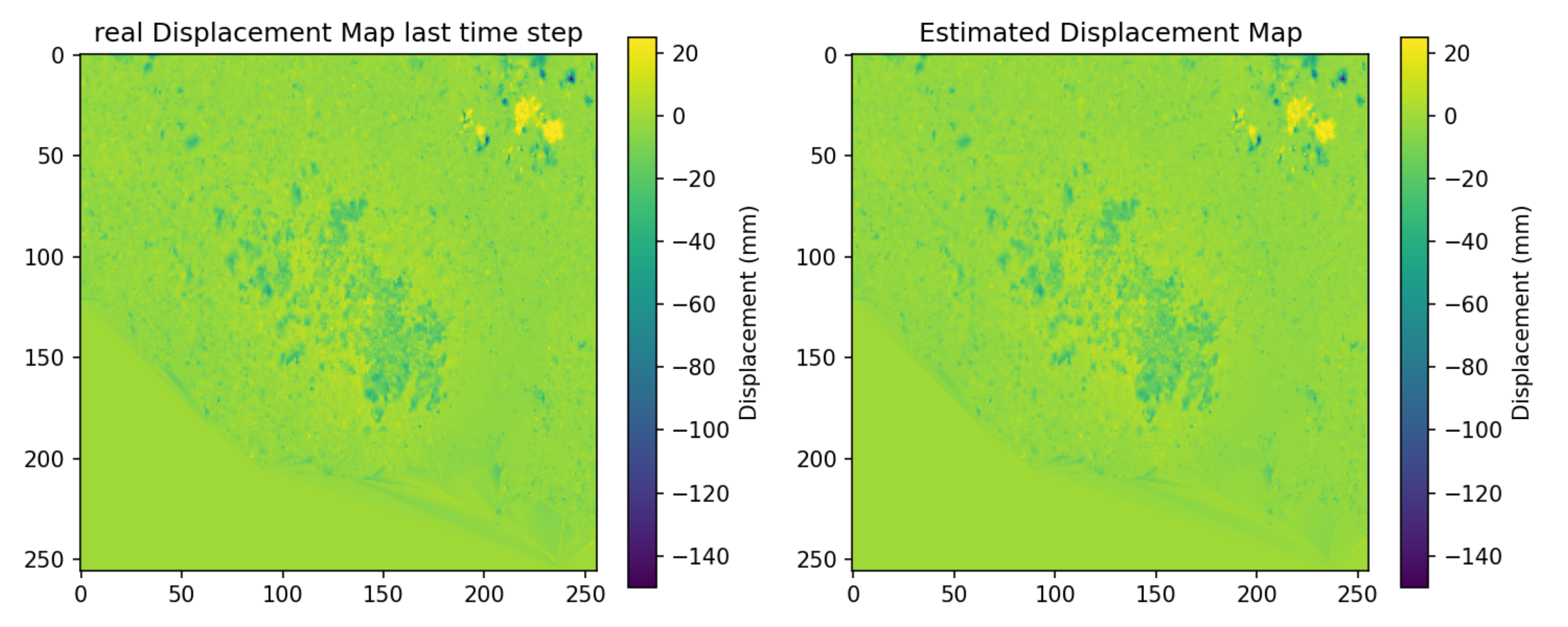}
    \caption{LASSO Regression Estimation}
    \label{fig:lasso_comp}
\end{subfigure}

\vspace{1em} % Adds vertical space between figures
% Subfigure for LightGBM
\begin{subfigure}{\textwidth}
    \centering
    \includegraphics[width=0.9\textwidth]{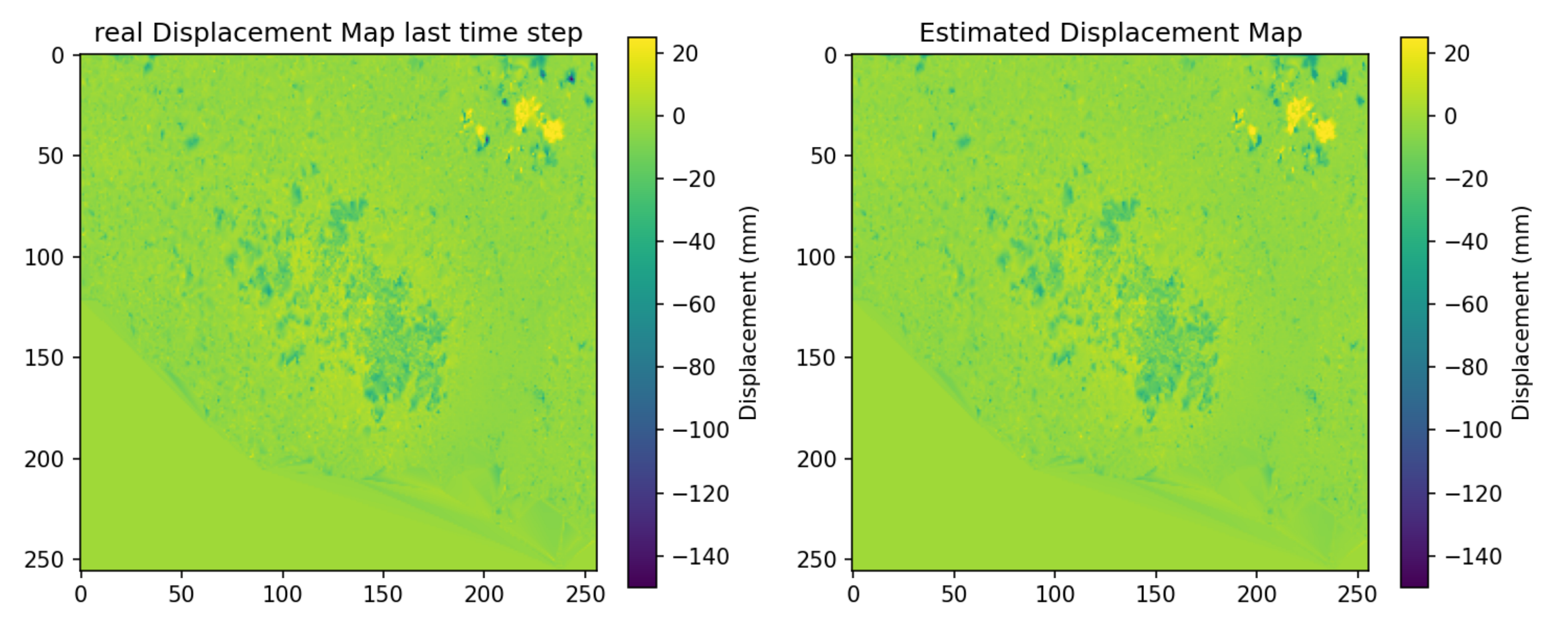}
    \caption{LightGBM Estimation}
    \label{fig:lgbm_comp}
\end{subfigure}

\vspace{1em} % Adds vertical space between figures

% Subfigure for CNN-LSTM
\begin{subfigure}{\textwidth}
    \centering
    \includegraphics[width=0.9\textwidth]{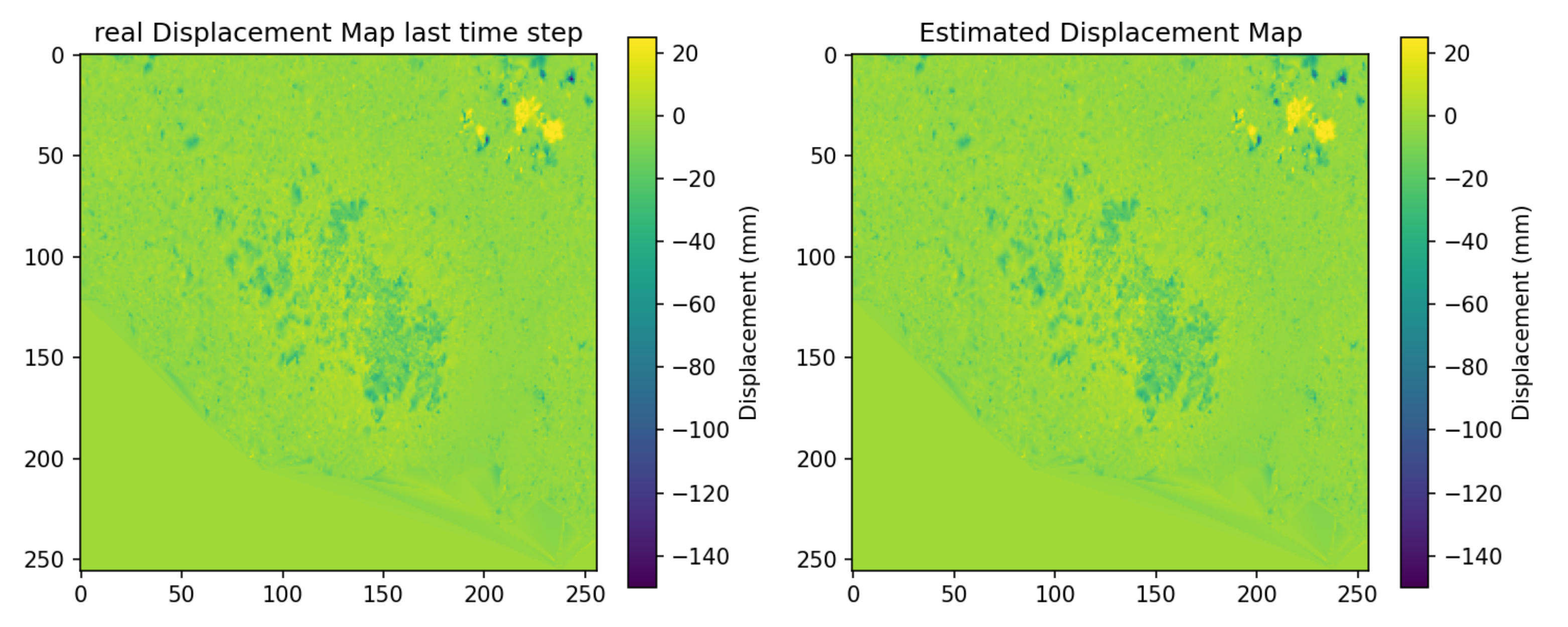}
    \caption{CNN-LSTM Estimation}
    \label{fig:cnnlstm_comp}
    
\end{subfigure}

\caption{Comparison between the ground truth displacement map (left side of each subfigure) and the estimated displacement maps (right side of each subfigure) from the three models. The models were trained on previous displacement data to predict the map for the final timestep. The results shown are for (a) LASSO Regression, (b) LightGBM, and (c) CNN-LSTM.}
\label{fig:sp-perfect-comparison}
\end{figure*}

Beyond quantitative metrics, a qualitative visual analysis of the predicted displacement maps provides critical insights into the performance of each model. This comparison allows us to assess how well each model captures the spatial heterogeneity and key deformation features present in the target data. Figure.~\ref{fig:sp-perfect-comparison} presents a side-by-side comparison of the real displacement map from the final time step against the estimated maps generated by the LASSO, LightGBM, and proposed CNN-LSTM models.

% paste 3 images in there. place on top, use full page inside of half opf the page. the images are: lasso,png, sp-perfect.png and frame_comparison.png. That is the comparison between estimated displacement map and real displacement map. We use the locations of the measurement points and their displacement value on last time step. and make models training with previous displacements to estimate the dispalcement map on last timestep. The image on the left is the real one. and the right one is models' estimation. the results from left to right are : lasso regression, cnn-lstm and lightgbm. Use this to introduce the image.
As illustrated in Figure.~\ref{fig:lasso_comp}, the LASSO regression model, which is a linear model, struggled significantly to replicate the complex spatial distribution of ground displacement. The resulting map is too smooth and does not capture the distinct boundaries of the central deformation zone. Furthermore, localized anomalies, such as the area of higher displacement in the top-right corner, are entirely absent from the prediction. This indicates that the LASSO model primarily captures the large-scale average displacement but is incapable of modeling the intricate, nonlinear spatial dependencies inherent in the data.

The LightGBM model, shown in Figure.~\ref{fig:lgbm_comp}, offers a marked improvement over the LASSO baseline. Identifies and delineates the primary area of displacement in the center of the map successfully. The general shape and magnitude of this zone are reasonably well-approximated. However, the model still exhibits a degree of spatial smoothing. The fine textures within the deformation zone are blurred, and the intensity of the peak displacement in the top right corner is underestimated compared to the ground truth. While LightGBM can model non-linear relationships, its pixel-independent approach limits its ability to learn from spatial context, leading to a loss of high-frequency spatial details.

In contrast, the prediction of the proposed CNN-LSTM model, presented in Figure .~\ref{fig:cnnlstm_comp}, demonstrates superior performance with remarkable visual fidelity to the real displacement map.  CNN-LSTM accurately reproduces sharp boundaries, internal textures, and the overall shape of the central deformation area. Crucially, it also captures the localized, high-magnitude displacement feature in the top-right corner with high precision in terms of both location and intensity. This high-quality reconstruction underscores the strength of the CNN-LSTM architecture. The convolutional layers effectively learn the spatial context and patterns of deformation at multiple scales, while the LSTM component models their evolution over time. This combined spatio-temporal learning approach enables the model to generate forecasts that are not only quantitatively accurate, but also spatially coherent and realistic.

\subsection{SHAP analysis on the prediction parameters}
As illustrated in Section.~\ref{SHAPIntro}, we applied a SHAP analysis after training for lightgbm model to help us understand how training variables affect the estimation. The result of the SHAP summary plot is shown in Figure.~\ref{fig:shap_summary}, and the result of the SHAP force plot is shown in Figure.~\ref{shap-force-pixel0}.

\subsubsection{Global Feature Importance}
\begin{figure*}[htbp]
    \centering
    \includegraphics[scale=0.8]{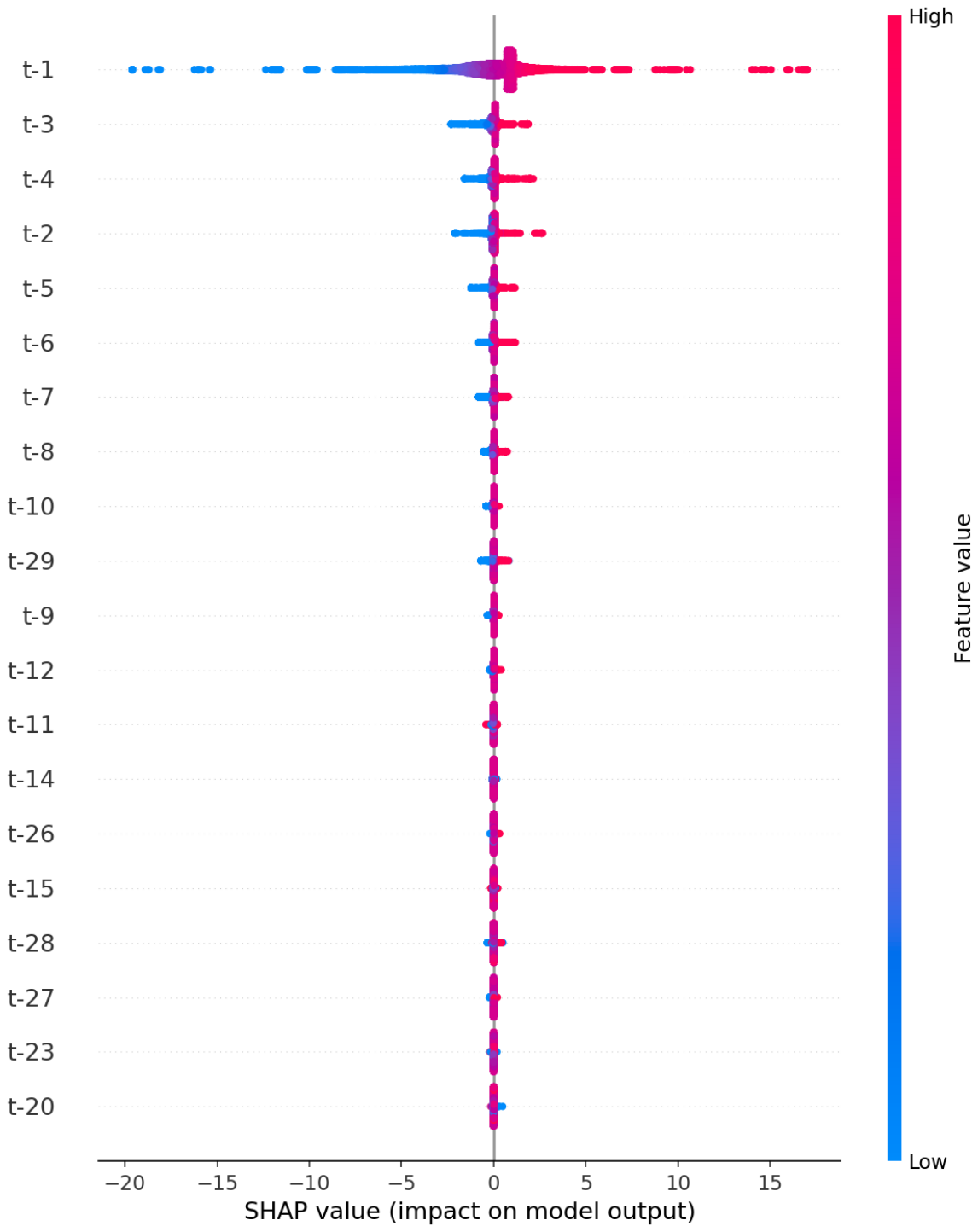}
    \caption{SHAP summary plot for the LightGBM model. This plot shows the distribution of SHAP values for each feature, ranked by importance. The most recent timestep, `t-1`, has the most significant impact on the model's predictions.}
    \label{fig:shap_summary}
\end{figure*}

Figure.~\ref{fig:shap_summary} shows the SHAP summary plot, which illustrates the impact of each characteristic in all test samples. The features are ranked by their global importance, with each point on a feature's row representing the SHAP value for a single-pixel's prediction.

The analysis reveals that the model predictions are overwhelmingly dominated by the most recent time step, `t-1`. This feature exhibits the largest spread of SHAP values, indicating its significant influence. The plot clearly shows a strong positive correlation: high displacement values at `t-1` (red points) have large positive SHAP values, pushing the model's prediction higher, while low values (blue points) have large negative SHAP values, driving the prediction lower.

The influence of other time steps diminishes rapidly. While `t-2` and `t-3` show some minor contribution, the impact of all older time steps is negligible, with their SHAP values clustered tightly around zero. This indicates that the LightGBM model has primarily learned a strong persistence or auto-regressive pattern, where the state of the system at the immediately preceding time step is the most critical factor for forecasting the future state.

\subsubsection{Local Prediction Analysis}
\begin{figure*}[t]
    \centerline{\includegraphics[scale=0.47]{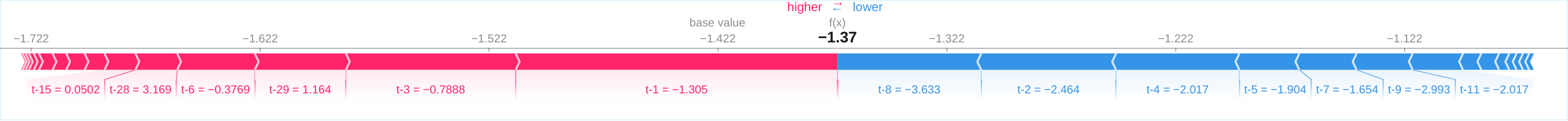}}
    \caption{SHAP force plot showing the feature contributions for a single prediction (pixel 0). Features in red increase the prediction, while features in blue decrease it. The final prediction is the balance point of these forces.}
    \label{shap-force-pixel0}
\end{figure*}

While the summary plot provides a global overview, a SHAP force plot allows us to decompose an individual prediction. Figure.~\ref{shap-force-pixel0} illustrates this for a single representative pixel of the test set. The plot shows how different feature values push the prediction away from the base value (the average prediction over the dataset, which is -2.768).

For this specific pixel, the model output is -2.546. The features shown in red, such as `t-1`, `t-293`, and `t-40`, have a positive SHAP value and contribute to increasing the prediction above the base value. In contrast, blue characteristics, such as `t-296` and `t-297`, have a negative impact, pushing the prediction lower. The final output value represents the balance of these competing forces. This local analysis confirms the global finding that the most recent time steps have the largest quantitative impact on individual predictions.

\section{Conclusions and Future Work}
In this study, we developed and evaluated a novel spatio-temporal prediction framework for forecasting ground displacement using Sentinel-1 InSAR data. By transforming sparse point measurements into a dense sequence of displacement maps, we enabled the application of advanced deep learning models capable of understanding both spatial and temporal contexts. We compared the performance of our proposed CNN-LSTM model against robust LightGBM and LASSO regression baselines.

Our results conclusively demonstrate that the CNN-LSTM architecture provides the most accurate predictions, achieving an R² value of 0.9901 and an RMSE of 0.2854. This superior performance underscores the importance of explicitly modeling the intertwined spatial and temporal characteristics inherent in ground deformation data. While the LASSO regression also provided a reasonable baseline, the LightGBM model, despite its power in many regression tasks, struggled to generalize, producing predictions with significant artifacts. Our SHAP analysis of the LightGBM model revealed that it learned a simple auto-regressive dependency on the most recent time step, failing to integrate information from the broader temporal sequence, which likely contributed to its higher error rate. The SHAP analysis reveals that the LightGBM model, despite its complexity, defaults to a simplistic persistence model, overwhelmingly relying on the t-1 feature. This failure to integrate longer-term temporal information and spatial context explains its poor performance. In contrast, the CNN-LSTM architecture is explicitly designed to capture both deep temporal dependencies via its LSTM component and spatial patterns via its CNN layers, which elucidates its vastly superior predictive accuracy. The primary contribution of this work is the validation of a deep learning pipeline that can produce high-fidelity, spatially coherent forecasts of ground movement, offering a valuable tool for infrastructure monitoring and geological risk assessment.

For future work, several avenues for enhancement can be explored. First, more advanced deep learning architectures, such as Vision Transformers or Graph Neural Networks (GNNs), could be investigated to potentially capture long-range spatial dependencies more effectively than CNNs. Second, the fusion of additional data sources—such as meteorological records (e.g., rainfall), geological surveys (e.g., soil type), and anthropogenic activity (e.g., construction schedules)—could provide valuable contextual information and further improve model accuracy. Finally, extending the model's predictive capabilities from a single-step to a multi-step forecast would significantly increase its practical utility for long-term planning and early-warning systems.

\section*{Acknowledgment}
This research was conducted with the financial support of Science Foundation Ireland under Grant Agreement No.\ 13/RC/2106\_P2 at the ADAPT SFI Research Centre at University College Dublin. ADAPT, the SFI Research Centre for AI-Driven Digital Content Technology is funded by Science Foundation Ireland through the SFI Research Centres Programme. This work is partly supported by China Scholarship Council (202306540013).

\bibliographystyle{IEEEtran}
\bibliography{./ref/longforms,./ref/references}

\begin{thebibliography}{10}
\providecommand{\url}[1]{#1}
\csname url@samestyle\endcsname
\providecommand{\newblock}{\relax}
\providecommand{\bibinfo}[2]{#2}
\providecommand{\BIBentrySTDinterwordspacing}{\spaceskip=0pt\relax}
\providecommand{\BIBentryALTinterwordstretchfactor}{4}
\providecommand{\BIBentryALTinterwordspacing}{\spaceskip=\fontdimen2\font plus
\BIBentryALTinterwordstretchfactor\fontdimen3\font minus \fontdimen4\font\relax}
\providecommand{\BIBforeignlanguage}[2]{{%
\expandafter\ifx\csname l@#1\endcsname\relax
\typeout{** WARNING: IEEEtran.bst: No hyphenation pattern has been}%
\typeout{** loaded for the language `#1'. Using the pattern for}%
\typeout{** the default language instead.}%
\else
\language=\csname l@#1\endcsname
\fi
#2}}
\providecommand{\BIBdecl}{\relax}
\BIBdecl

\bibitem{GroundWater}
T.~Q. Cuong, D.~H. Tong~Minh, L.~Van~Trung, and T.~Le~Toan, ``Ground subsidence monitoring in vietnam by multi-temporal insar technique,'' in \emph{2015 IEEE International Geoscience and Remote Sensing Symposium (IGARSS)}, 2015, pp. 3540--3543.

\bibitem{DisAndLandS}
L.~Zou, K.~Takahashi, and M.~Sato, ``Displacement estimation and monitoring from ground-based sar amplitude components,'' in \emph{2014 Asia-Pacific Microwave Conference}, 2014, pp. 1327--1329.

\bibitem{landDis}
Y.~Cheng, W.~Zhang, F.~Zhao, and S.~Yang, ``Typical landslide monitoring early warning and susceptibility evaluation based on bp neural network model in zhouzhi county,'' in \emph{2024 IEEE 2nd International Conference on Sensors, Electronics and Computer Engineering (ICSECE)}, 2024, pp. 1--6.

\bibitem{UrbanDef1}
D.~Tapete, F.~Cigna, R.~Lasaponara, N.~Masini, and P.~Milillo, ``Deformation analysis of a metropolis from c- to x-band psi: Proof-of-concept with cosmo-skymed over rome, italy,'' in \emph{2015 IEEE International Geoscience and Remote Sensing Symposium (IGARSS)}, 2015, pp. 4606--4609.

\bibitem{UrbanDef2}
F.~Zhao, J.~J. Mallorqui, and J.~M. Lopez-Sanchez, ``Impact of sar image resolution on polarimetric persistent scatterer interferometry with amplitude dispersion optimization,'' \emph{IEEE Transactions on Geoscience and Remote Sensing}, vol.~60, pp. 1--10, 2022.

\bibitem{UrbanDef3}
\BIBentryALTinterwordspacing
S.~Azadnejad, A.~Hrysiewicz, A.~Trafford, F.~O'Loughlin, E.~Holohan, F.~Kelly, and S.~Donohue, ``Insar supported by geophysical and geotechnical information constrains two-dimensional motion of a railway embankment constructed on peat,'' \emph{Engineering Geology}, vol. 333, p. 107493, 2024. [Online]. Available: \url{https://www.sciencedirect.com/science/article/pii/S0013795224000917}
\BIBentrySTDinterwordspacing

\bibitem{OSMANOGLU201690}
\BIBentryALTinterwordspacing
B.~Osmanoğlu, F.~Sunar, S.~Wdowinski, and E.~Cabral-Cano, ``Time series analysis of insar data: Methods and trends,'' \emph{ISPRS Journal of Photogrammetry and Remote Sensing}, vol. 115, pp. 90--102, 2016, theme issue 'State-of-the-art in photogrammetry, remote sensing and spatial information science'. [Online]. Available: \url{https://www.sciencedirect.com/science/article/pii/S0924271615002269}
\BIBentrySTDinterwordspacing

\bibitem{EGMSDescript}
R.~Palamà, M.~Cuevas-González, A.~Barra, Q.~Gao, S.~Shahbazi, J.~A. Navarro, O.~Monserrat, and M.~Crosetto, ``Automatic ground deformation detection from european ground motion service products,'' in \emph{IGARSS 2023 - 2023 IEEE International Geoscience and Remote Sensing Symposium}, 2023, pp. 8190--8193.

\bibitem{AZADNEJAD2020101950}
\BIBentryALTinterwordspacing
S.~Azadnejad, Y.~Maghsoudi, and D.~Perissin, ``Evaluation of polarimetric capabilities of dual polarized sentinel-1 and terrasar-x data to improve the psinsar algorithm using amplitude dispersion index optimization,'' \emph{International Journal of Applied Earth Observation and Geoinformation}, vol.~84, p. 101950, 2020. [Online]. Available: \url{https://www.sciencedirect.com/science/article/pii/S0303243419302995}
\BIBentrySTDinterwordspacing

\bibitem{EuroSATYao}
W.~Yao, S.~Donohue, and S.~Dev, ``Optimizing remote sensing image classification with spectral indices and convolutional neural networks,'' in \emph{2023 IEEE 7th Conference on Energy Internet and Energy System Integration (EI2)}, 2023, pp. 3628--3633.

\bibitem{AZEEM2024100515}
\BIBentryALTinterwordspacing
M.~A. Azeem and S.~Dev, ``A performance and interpretability assessment of machine learning models for rainfall prediction in the republic of ireland,'' \emph{Decision Analytics Journal}, vol.~12, p. 100515, 2024. [Online]. Available: \url{https://www.sciencedirect.com/science/article/pii/S277266222400119X}
\BIBentrySTDinterwordspacing

\bibitem{InSARLS1}
A.~Shakeel, R.~J. Walters, S.~K. Ebmeier, and N.~A. Moubayed, ``Aladdin: Autoencoder-lstm-based anomaly detector of deformation in insar,'' \emph{IEEE Transactions on Geoscience and Remote Sensing}, vol.~60, pp. 1--12, 2022.

\bibitem{InSARLS2}
Y.~Wen, X.~Wan, D.~Yuan, L.~Zhang, D.~Ge, and L.~Zhang, ``C-lstm for mt-insar ground deformation prediction,'' in \emph{IGARSS 2024 - 2024 IEEE International Geoscience and Remote Sensing Symposium}, 2024, pp. 11\,016--11\,019.

\bibitem{WesIe}
W.~Yao, S.~Azadnejad, S.~Donohue, and S.~Dev, ``A study on spatial prediction models and influential parameters of deformation in western ireland,'' in \emph{2024 17th International Congress on Image and Signal Processing, BioMedical Engineering and Informatics (CISP-BMEI)}, 2024, pp. 1--6.

\bibitem{EGMS}
M.~Costantini, F.~Minati, F.~Trillo, A.~Ferretti, F.~Novali, E.~Passera, J.~Dehls, Y.~Larsen, P.~Marinkovic, M.~Eineder, R.~Brcic, R.~Siegmund, P.~Kotzerke, M.~Probeck, A.~Kenyeres, S.~Proietti, L.~Solari, and H.~S. Andersen, ``European ground motion service (egms),'' in \emph{2021 IEEE International Geoscience and Remote Sensing Symposium IGARSS}, 2021, pp. 3293--3296.

\bibitem{LightGBM}
\BIBentryALTinterwordspacing
B.~Li, K.~Liu, M.~Wang, Y.~Wang, Q.~He, L.~Zhuang, and W.~Zhu, ``High-spatiotemporal-resolution dynamic water monitoring using lightgbm model and sentinel-2 msi data,'' \emph{International Journal of Applied Earth Observation and Geoinformation}, vol. 118, p. 103278, 2023. [Online]. Available: \url{https://www.sciencedirect.com/science/article/pii/S1569843223001000}
\BIBentrySTDinterwordspacing

\bibitem{LSTMorigin}
\BIBentryALTinterwordspacing
S.~Hochreiter and J.~Schmidhuber, ``Long short-term memory,'' \emph{Neural Comput.}, vol.~9, no.~8, p. 1735–1780, Nov. 1997. [Online]. Available: \url{https://doi.org/10.1162/neco.1997.9.8.1735}
\BIBentrySTDinterwordspacing

\bibitem{Lasso}
J.~Ranstam and J.~A. Cook, ``Lasso regression,'' \emph{Journal of British Surgery}, vol. 105, no.~10, pp. 1348--1348, 2018.

\bibitem{MSE}
K.~Venkatraman, M.~Akashvarma, and S.~Siddharth, ``Enhancing software test effort estimation using ensemble learning algorithms,'' in \emph{2024 4th International Conference on Intelligent Technologies (CONIT)}, 2024, pp. 1--5.

\bibitem{lundberg2017unified}
S.~M. Lundberg and S.-I. Lee, ``A unified approach to interpreting model predictions,'' \emph{Advances in neural information processing systems}, vol.~30, 2017.

\end{thebibliography}

\end{document}